\documentclass{article}
\usepackage{arxiv}
\usepackage[utf8]{inputenc} 
\usepackage[T1]{fontenc}    
\usepackage{hyperref}       
\usepackage{url}            
\usepackage{booktabs}       
\usepackage{amsfonts}       
\usepackage{nicefrac}       
\usepackage{microtype}      
\usepackage{lipsum}
\usepackage[authoryear]{natbib}
\usepackage{subcaption}
\usepackage{graphicx}
\graphicspath{ {./figures/} }

\title{Enhancing Lithological Mapping with Spatially Constrained Bayesian Network (SCB-Net): An Approach for Field Data-Constrained Predictions with Uncertainty Evaluation}

\author{
 Victor Silva dos Santos \\
  Eau Terre Environnement (ETE)\\
  Institut National de la Recherche Scientifique (INRS)\\
  490 Couronne St, Quebec City, Quebec G1K 9A9 \\
  \texttt{victor.santos@inrs.ca} \\
   \And
Erwan Gloaguen \\
  Eau Terre Environnement (ETE)\\
  Institut National de la Recherche Scientifique (INRS)\\
  490 Couronne St, Quebec City, Quebec G1K 9A9 \\
  \texttt{erwan.gloaguen@inrs.ca} \\
  \And
 Shiva Tirdad \\
  Geological Survey of Canada (GSC)\\
  Natural Resources Canada (NRCan)\\
  490 Couronne St, Quebec City, Quebec G1K 9A9 \\
  \texttt{shiva.tirdad@nrcan-rncan.gc.ca} \\
}

\begin{document}
\maketitle
\begin{abstract}
  Geological maps are an extremely valuable source of information for the Earth sciences. They provide insights into mineral exploration, vulnerability to natural hazards, and many other applications. These maps are created using numerical or conceptual models that use geological observations to extrapolate data. Geostatistical techniques have traditionally been used to generate reliable predictions that take into account the spatial patterns inherent in the data. However, as the number of auxiliary variables increases, these methods become more labor-intensive. Additionally, traditional machine learning methods often struggle with spatially correlated data and extracting valuable non-linear information from geoscientific datasets. 
  To address these limitations, a new architecture called the Spatially Constrained Bayesian Network (SCB-Net) has been developed. The SCB-Net aims to effectively exploit the information from auxiliary variables while producing spatially constrained predictions. It is made up of two parts, the first part focuses on learning underlying patterns in the auxiliary variables while the second part integrates ground-truth data and the learned embeddings from the first part. Additionally, to assess model uncertainty, a technique called Monte Carlo dropout is used as a Bayesian approximation. 
  The SCB-Net has been applied to two selected areas in northern Quebec, Canada, and has demonstrated its potential in generating field-data-constrained lithological maps while allowing assessment of prediction uncertainty for decision-making. This study highlights the promising advancements of deep neural networks in geostatistics, particularly in handling complex spatial feature learning tasks, leading to improved spatial information techniques.
\end{abstract}


\section{Introduction}
Geological maps are essential sources of information in the Earth sciences. They offer valuable insights into mineral exploration, environmental management, and hazard assessment. The creation of geological maps involves the use of conceptual or numerical models, which are used to extrapolate geological data to locations where the ground truth is unknown. This process entails identifying and characterizing various rock types, minerals, and geological structures. To predict the spatial distribution of rock units based on auxiliary information, different methods have been employed, including geostatistical techniques and machine learning algorithms. Geostatistical techniques like indicator cokriging are commonly used to generate field-data-constrained estimates \citep{wang2013predictive, macikag2020application, guartan2021predictive}. However, with an increase in the number of auxiliary variables, these techniques tend to become time-consuming \citep{cuba2009selection}, and their inherent linearity assumption between variables limits their capability \citep{kirkwood2022bayesian}.

While machine learning and deep learning algorithms have shown success in predictive lithological mapping (PLM) \citep{cracknell2014geological, harris2015predictive, costa2019predictive, wang2021lithological, cedou2022preliminary}, most published research on PLM focuses on validating preexisting geological maps \citep{cracknell2014geological, costa2019predictive, shebl2021lithological, wang2021lithological}. These studies use geological maps as the target for machine learning models, allowing them to identify discrepancies between the predicted and actual geological maps. However, in regional mapping scenarios, geological maps are often outdated and lack the necessary level of detail to serve as accurate ground-truth references. Incorporating field data becomes crucial in such cases to generate reliable predictions. Despite this need for field data, there is a lack of research proposing solutions to create maps that incorporate field data and allow for the quantification of uncertainties in predicted locations.

U-net is a type of Convolutional Neural Network (CNN) introduced by \citep{ronneberger2015u}. It was originally designed for medical image segmentation tasks, where the goal is to partition an input image into meaningful segments or regions of interest. Analogously, geologists divide study areas into distinct sub-regions when interpreting possible geological contacts using remote sensing (RS) imagery. Notably, in traditional geological mapping, final maps must honor the field observations, which are typically sparse and irregularly distributed. Nonetheless, within the domain of machine learning, this is generally associated to over-fitting, which prevents the model from generalizing to previously unseen data \citep{dietterich1995overfitting, ying2019overview}. 

Probabilistic modeling has gained increasing importance in the field of machine learning. These models are rooted in probability theory and offer a means to handle uncertainty within data \citep{ghahramani2015probabilistic, murphy2022probabilistic}. In contrast to conventional machine learning models, which provide a single prediction or decision, probabilistic models yield a distribution representing various potential predictions or decisions. This distribution serves to quantify uncertainty and enable more informed decision-making \citep{murphy2022probabilistic}. \cite{gal2016dropout} introduced a method for approximating Bayesian inference in deep neural networks using a technique called Monte Carlo dropout. This technique allows the model to learn a distribution over the weights, which can be used to estimate the uncertainty in the predictions. Subsequently, \cite{kirkwood2022bayesian} showcased the potential of Bayesian deep learning in geosciences, particularly in the context of spatial interpolation when auxiliary information, such as satellite imagery, is available.

In our research, we have presented a new method that combines deep learning techniques with Bayesian inference for PLM. This approach, called Spatially Constrained Bayesian Network (SCB-Net), makes it possible to generate accurate predictions while also taking into account uncertainty in predicted locations, particularly in scenarios where auxiliary information is available. To achieve this, we have used U-net as the fundamental building block of our architecture and developed a custom loss function that addresses the challenges associated with class imbalance and over-smoothing.  We have demonstrated the effectiveness of our approach by conducting a case study in two areas located in the North of Quebec, Canada. As part of this study, we have integrated a range of remotely sensed data sources, including multi-spectral and RADAR imagery, magnetic, and terrain elevation data, as auxiliary information.

\section{Methods}

\subsection{Building block model: U-net}

The U-Net architecture is characterized by its U-shaped structure, which consists of an encoder path and a decoder path. The encoder path performs a series of down-sampling operations to capture high-level features and semantic information from the input image. On the other hand, the decoder path involves up-sampling and concatenation of feature maps from the encoder path to produce a high-resolution segmentation map. The skip connections between the encoder and decoder paths allow the U-Net to retain fine-grained spatial details, effectively addressing the problem of information loss during down-sampling. The U-Net has gained widespread popularity due to its remarkable ability to achieve accurate and precise image segmentation results, making it a valuable tool in various medical image analysis tasks \citep{ronneberger2015u}.
      
\subsection{Proposed architecture}

\begin{figure}
  \centering
  \includegraphics[width=0.8\textwidth]{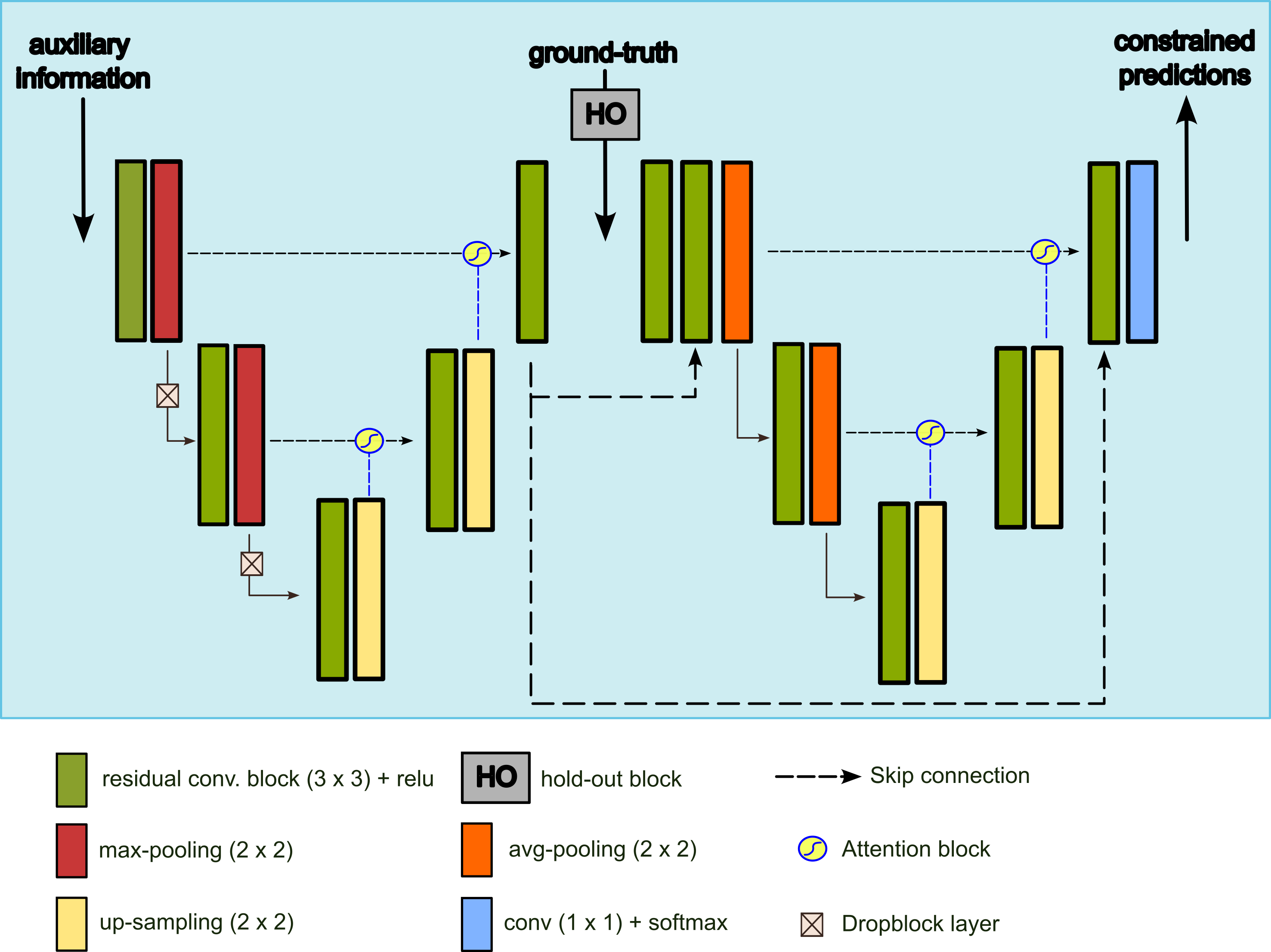}
  \caption{Our proposed architecture for obtaining field-data constrained predictions. The model is basically composed of two Attention Res-Unets, which are responsible for extracting features from the auxiliary data and fusing them with the sparse probability masks.}
  \label{fig:ntw_arch}
\end{figure}

Our approach combines two Attention Res-Unets to leverage the power of deep learning in lithological mapping while effectively incorporating both auxiliary information and ground-truth geological data (Figure \ref{fig:ntw_arch}). It was inspired by the W-net, a model proposed by \citep{xia2017w} for unsupervised image segmentation tasks. In its first part, our model focuses on extracting meaningful features from the input images, capturing important spatial relationships and relevant patterns present in the remotely sensed data. Meanwhile, incorporating ground-truth geological data is essential for improving the accuracy and reliability of lithological mapping. To achieve this, in its second part, our model concatenates the extracted representations from the first part with the sparse probability masks. This allows the fusion of both sources of information and captures the intricacies between the primary and secondary variables. This ensures that the model honors the available field data while generating lithological predictions.

The role of residual blocks is to facilitate information flow within the network while addressing the vanishing gradient problem, especially in very deep networks \citep{he2016deep}. Meanwhile, attention blocks enable networks to focus on different parts of the input data by assigning weights to shallower representations based on deeper, more complex representations in the network, guiding the model on where to direct its attention \citep{oktay2018attention}. This capability is particularly advantageous in predictive mapping using remotely sensed data, where the model needs to extract multiple scale features from the input data without losing too much spatial information.

\subsection{Model training}

\begin{table}
  \centering
  \caption{Hyperparameters adopted for model training.}\label{hyperparameters}
  \vspace{0.5cm} 
  \begin{tabular}{c|c}
    \hline
    \textbf{Hyperparameter} & \textbf{Value}\\
    \hline
        Batch size & 16 \\
        Patch size & 160x160 \\
        Dropout rate & 0.3 \\
        Hold-out rate & 0.3 \\
        Early stopping ($\Delta$) & $10^{-3}$ \\
        Learning rate & $5 \times 10^{-5}$ \\
        Patience (epochs) & 50 \\
        Maximum epochs & 500 \\
  \end{tabular}
\end{table}

\begin{figure}
  \centering
  \includegraphics[width=0.75\textwidth]{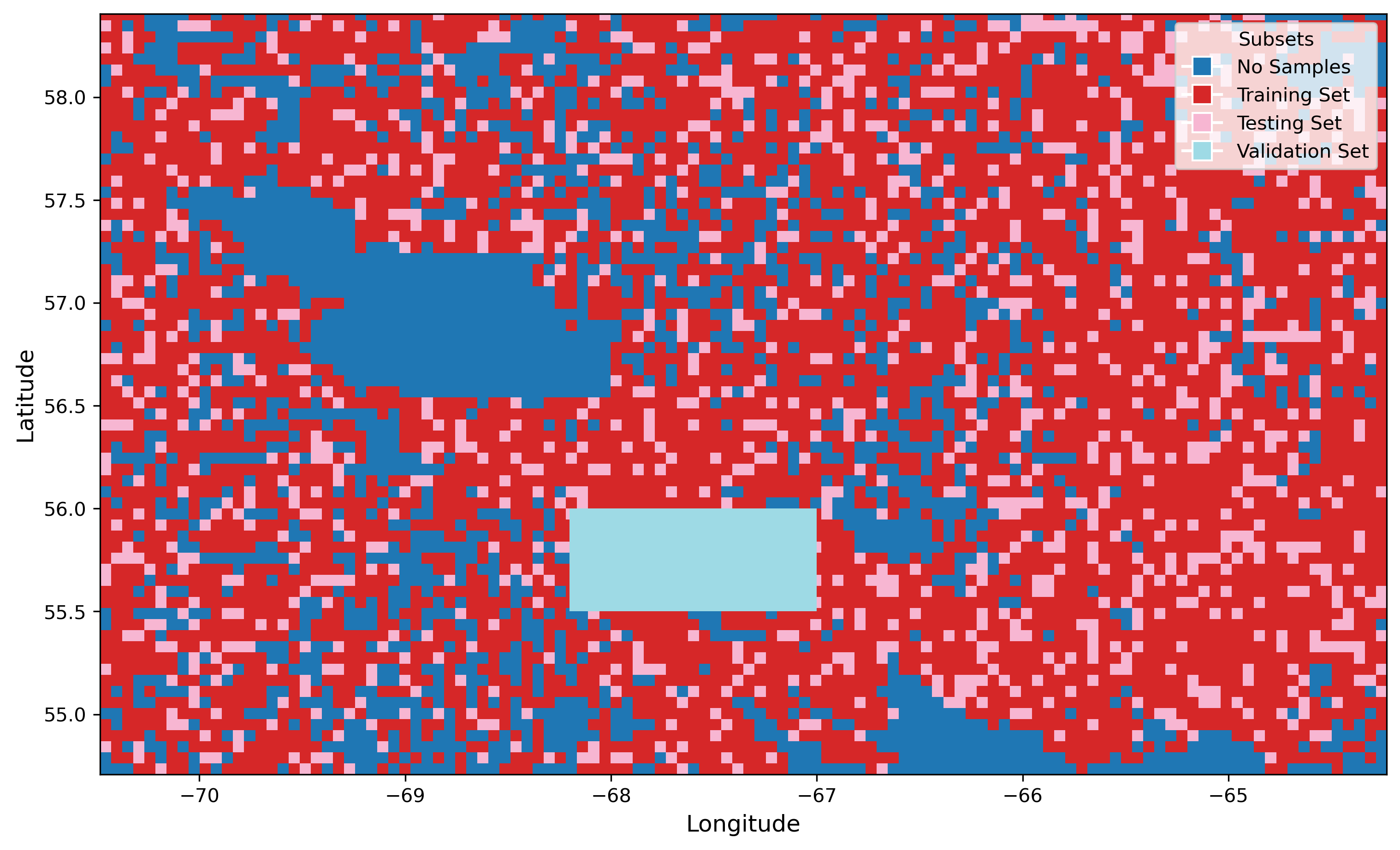}
  \caption{The northeast area is divided into spatial blocks, each measuring 15x15 pixels. The training set is represented by pink blocks, while the testing set is represented by orange blocks. The green zone is reserved for validation purposes. You may notice some dark-blue blocks which contain no field-samples.}
  \label{fig:spatial_blocks}
\end{figure}

The hyperparameters used to train the model are listed in Table \ref{hyperparameters}. We utilized the Adam optimizer with a learning rate of 0.0005 and a batch size of 16 for 234 epochs (we implemented early stopping based on testing accuracy). To account for class imbalance in our dataset, we weighted accuracy proportionally. The area was divided into spatial blocks of 15 by 15 pixels (Figure \ref{fig:spatial_blocks}), with an 80\% - 20\% split for training and testing, respectively. We also set aside a larger zone in the center of the study area for future validation (Figure \ref{fig:spatial_blocks}). During training, we employed a data hold-out strategy for cross-validation. At each epoch, 50\% of the training samples were randomly selected, and the integrity of the training samples was compared. To train the model, we generated 2,600 pairs of images-masks measuring 160x160 pixels, with up to 80\% overlap. For multi-scale analysis, we under-sampled the original images and masks by a factor of 2 and generated approximately 30\% of the tiles, resulting in a four times larger surface coverage. Additionally, about 25\% of the tiles were rotated from -12 to +12 degrees.

The model was trained on a single NVIDIA GeForce RTX 3070 GPU equipped with 8 GB of memory. We implemented our model using the TensorFlow framework \citep{abadi2016tensorflow}. Python language was used for all other tasks, including data preparation, model training, and evaluation.

\subsection{Custom loss function}

To enhance the performance of our model, we developed a specialized loss function that combines focal categorical cross-entropy with a dilation operator. This loss function addresses issues related to imbalanced classes and facilitates learning of long-range relationships within the auxiliary data.

The focal categorical cross-entropy (FCCE) loss is traditionally employed in addressing multiclass classification problems with unbalanced classes. FCCE quantifies the dissimilarity between predicted probabilities and ground-truth labels, with a focus on the most challenging examples within the training dataset. The objective is to minimize the equation (Eq. \ref{eq:fcce}):

\begin{equation}\label{eq:fcce}
  \mathcal{L}_{FCCE} = - \sum_{i=1}^{N} \sum_{j=1}^{C} y_{ij} (1 - p_{ij})^{\gamma} \log(p_{ij}),
\end{equation}
where $N$ is the number of samples, $C$ is the number of classes, $y_{ij}$ is the ground-truth label for sample $i$ and class $j$, $p_{ij}$ is the predicted probability for sample $i$ and class $j$, and $\gamma$ is the focusing parameter. The focusing parameter $\gamma$ is a tunable hyperparameter that controls the degree of focusing. When $\gamma = 0$, the FCCE loss is equivalent to the standard categorical cross-entropy loss. In our study, we set $\gamma = 2$.

To better capture the spatial dependencies between sampled locations and remotely sensed imagery, we incorporate a dilation operator into our loss function. Dilation is a fundamental morphological operation used in image processing to expand the boundaries of an object \citep{goyal2011morphological}. Since the target information is sparse, this operator enhances the model's ability to capture long-range relationships in the data, resulting in more realistic texture, especially in regions with few or no samples. In our study, we applied dilation to the predicted probabilities using filters of sizes 1x1 (i.e., no filter), 3x3, 5x5, and 11x11, comparing their dissimilarity to the ground-truth information at every step. Furthermore, different weights were assigned to each filter size (0.2, 0.3, 0.25, and 0.25, respectively).

\subsection{Uncertainty quantification}

As previously mentioned, we incorporate Monte Carlo dropout into our framework to enable the quantification of model uncertainty. This technique leverages multiple network configurations resulting from neuron dropout, where different sets of neurons are activated at each prediction, to evaluate model stability. Since predictions vary based on the set of neurons used during inference, Monte Carlo dropout serves as an approximation of the posterior distribution of the weights of a network \citep{gal2016dropout}. This distribution can be employed to estimate the uncertainty in predictions, and it has been successfully applied in spatial interpolation using neural networks, as demonstrated by \cite{kirkwood2022bayesian}.

In this study, we utilized Dropblock layers instead of classic Dropout layers. Dropblock is a regularization technique that eliminates contiguous regions of feature maps rather than individual neurons. This method has demonstrated greater effectiveness than Dropout in convolutional neural networks, particularly due to the spatially structured nature of images \citep{ghiasi2018dropblock}.

For the first Attention Res-Unet, we implemented Dropblock after each convolutional block, employing a block size of 5x5 and a drop rate of 0.3. Conversely, no Dropblock was applied to the second Attention Res-Unet to ensure minimal prediction variance at sampled locations.

\section{Study areas - North of Quebec}

\begin{figure}
  \centering
  \includegraphics[width=0.75\textwidth]{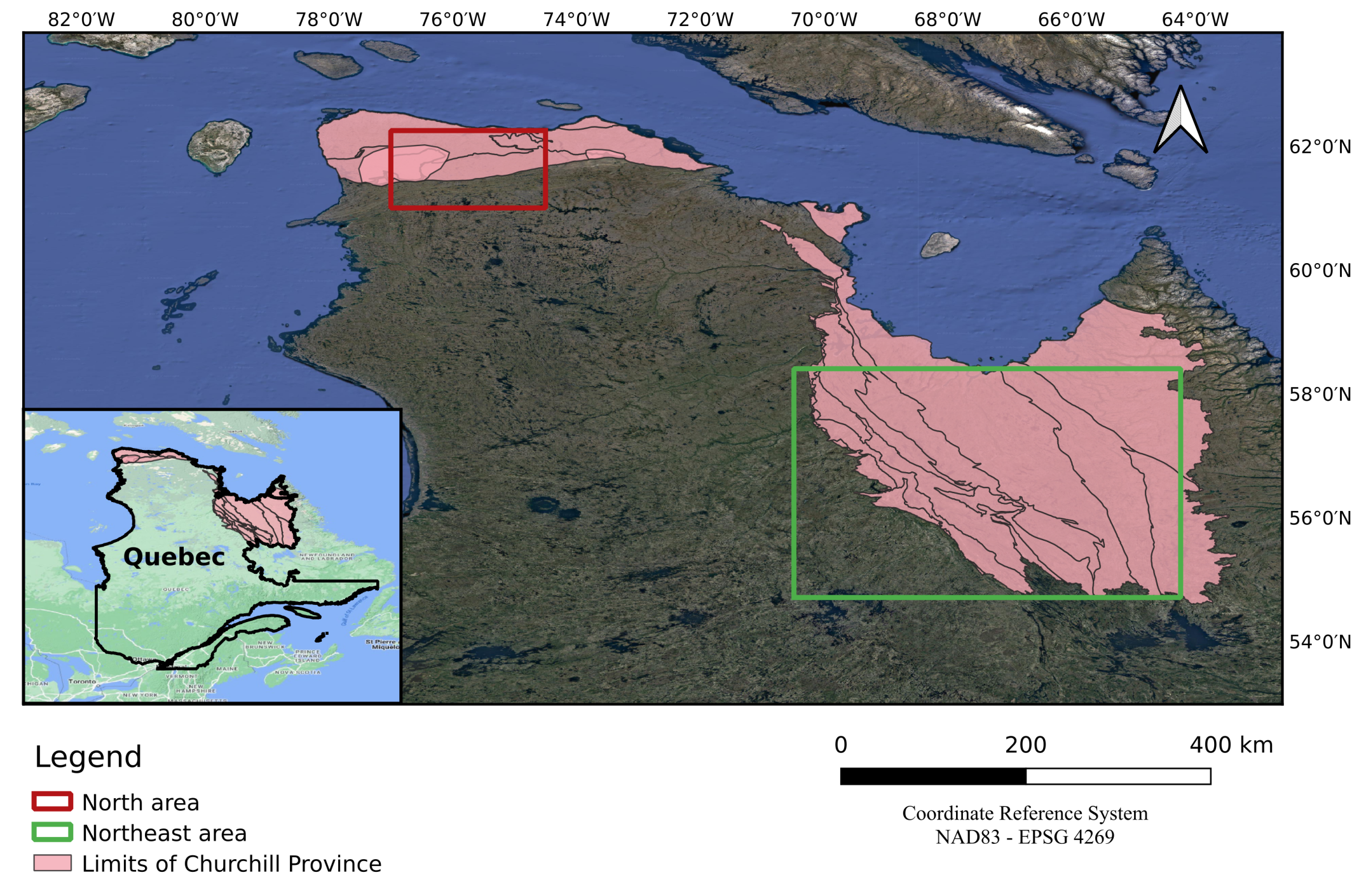}
  \caption{Study areas in the province of Quebec. The northeast area is highlighted in green, while the northern area is highlighted in red.}
  \label{fig:study_areas}
\end{figure}

For our research, we chose two locations in the northern part of Quebec Province, within the Churchill Province boundaries (Figure \ref{fig:study_areas}). These areas have a diverse range of lithologies, including granitoids, mafic and ultramafic plutonic rocks, mafic volcanic formations, metamorphic rocks, and sedimentary deposits. We selected these locations due to the availability of field data and auxiliary information, as well as their similar geology in terms of lithologies and tectonics \citep{mernchurchill2020}. Our primary focus is on the northeast area, while the northern area is used as a testbed for our transfer learning strategy.

\section{Data preparation}

\subsection{Remotely sensed data}

The satellite data was obtained from the Google Earth Engine (GEE) platform. It consists of multi-spectral imagery (Sentinel 2), RADAR imagery (Global PALSAR/PALSAR 2 Mosaic), and terrain elevation data (ALOS-PALSAR). To reduce the impact of cloud and snow presence in multi-spectral images, as well as noise in radar data, we calculated the median of the summer images collected between 2015 and 2021 from both sensors.

The magnetic data were downloaded from the Quebec government's spatially referenced geomineral information system (SIGÉOM). This dataset comprises the total residual magnetic field from a compilation of aeromagnetic surveys. These surveys were conducted with an 800-meter spacing between flight lines and at a flight height of 300 meters. 
    
\begin{figure}
  \centering
  \includegraphics[width=0.75\textwidth]{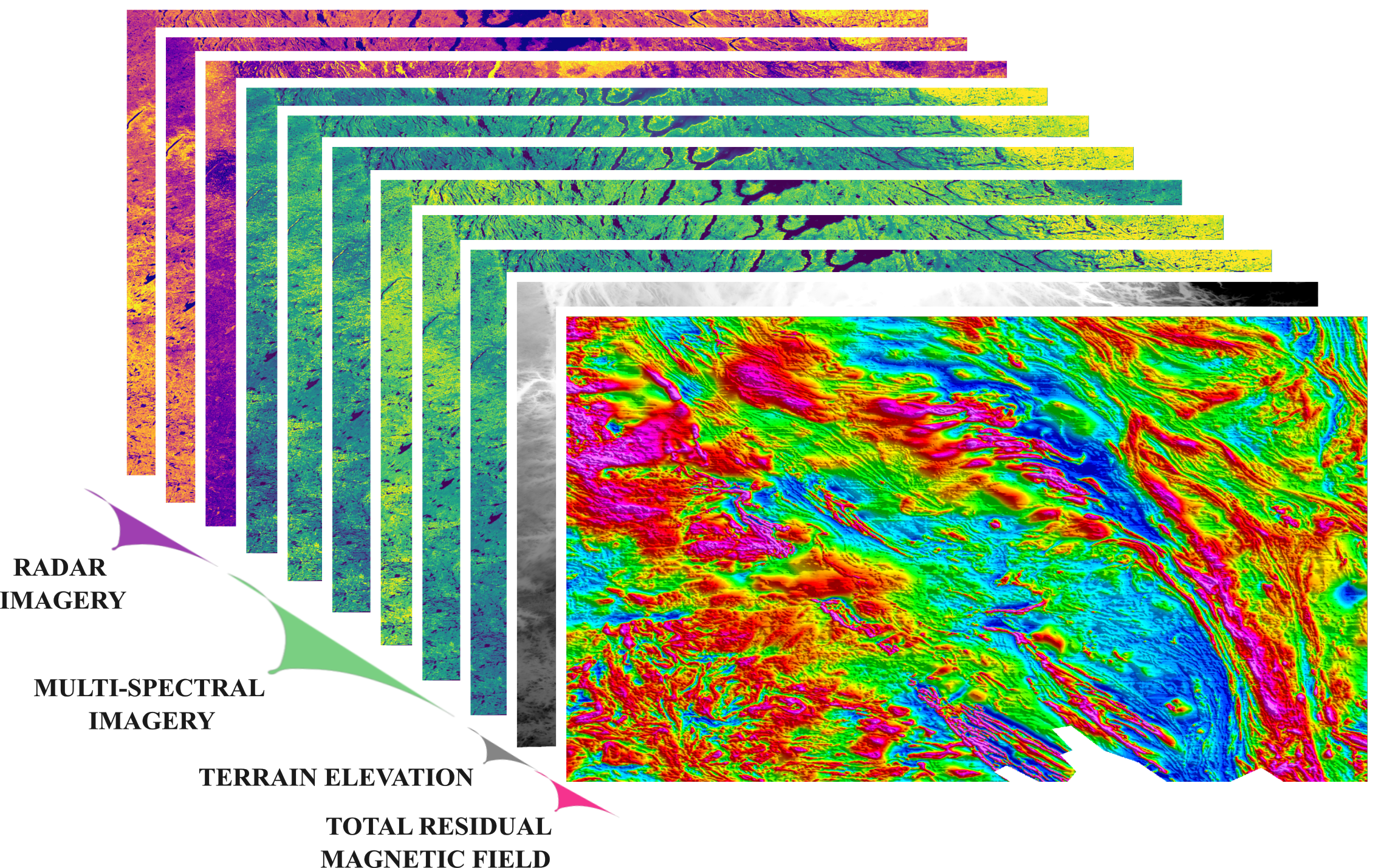}
  \caption{Remote sensing dataset used as auxiliary variables for lithology prediction.}
  \label{fig:rs_dataset}
  
\end{figure}

To ensure consistency, all remotely sensed data were re-interpolated to a reference grid with a pixel size of 400 meters. Figure \ref{fig:rs_dataset} displays the remotely sensed data used in this study for the northeast area, and a summary of the main characteristics of the data is presented in Table \ref{tb:summary_rs}.

\begin{table}
  \centering
  \caption{Summary of data used as auxiliary information for lithology classification.}
  \label{tb:summary_rs}
  \vspace{0.5cm} 
  \begin{tabular}{p{0.4\linewidth}|p{0.5\linewidth}}
    \hline
    \textbf{Description}\\
    \hline\\
    Multi-spectral Imagery & Red, Green, Blue, NIR, SWIR1, and SWIR2 bands - original pixel size varies from 10 to 20 meters (Sentinel 2)\\[0.75cm]
    RADAR Imagery & Polarizations HH, HV, and HH/HV - original pixel size is 25 meters (PALSAR/PALSAR 2)\\[0.75cm]
    Digital Elevation Model (DEM) & Elevation with radiometric terrain correction - original pixel size of 10 meters (ALOS-PALSAR)\\[0.75cm] 
    Magnetic data & Total residual magnetic field - 800-meter spacing between flight lines in average (Federal dataset, Canada)\\[0.75cm]  
  \end{tabular}
\end{table}

\subsection{Field data}

\begin{figure}
    \centering
    \begin{subfigure}[b]{0.70\linewidth}
        \centering
        \includegraphics[width=\linewidth]{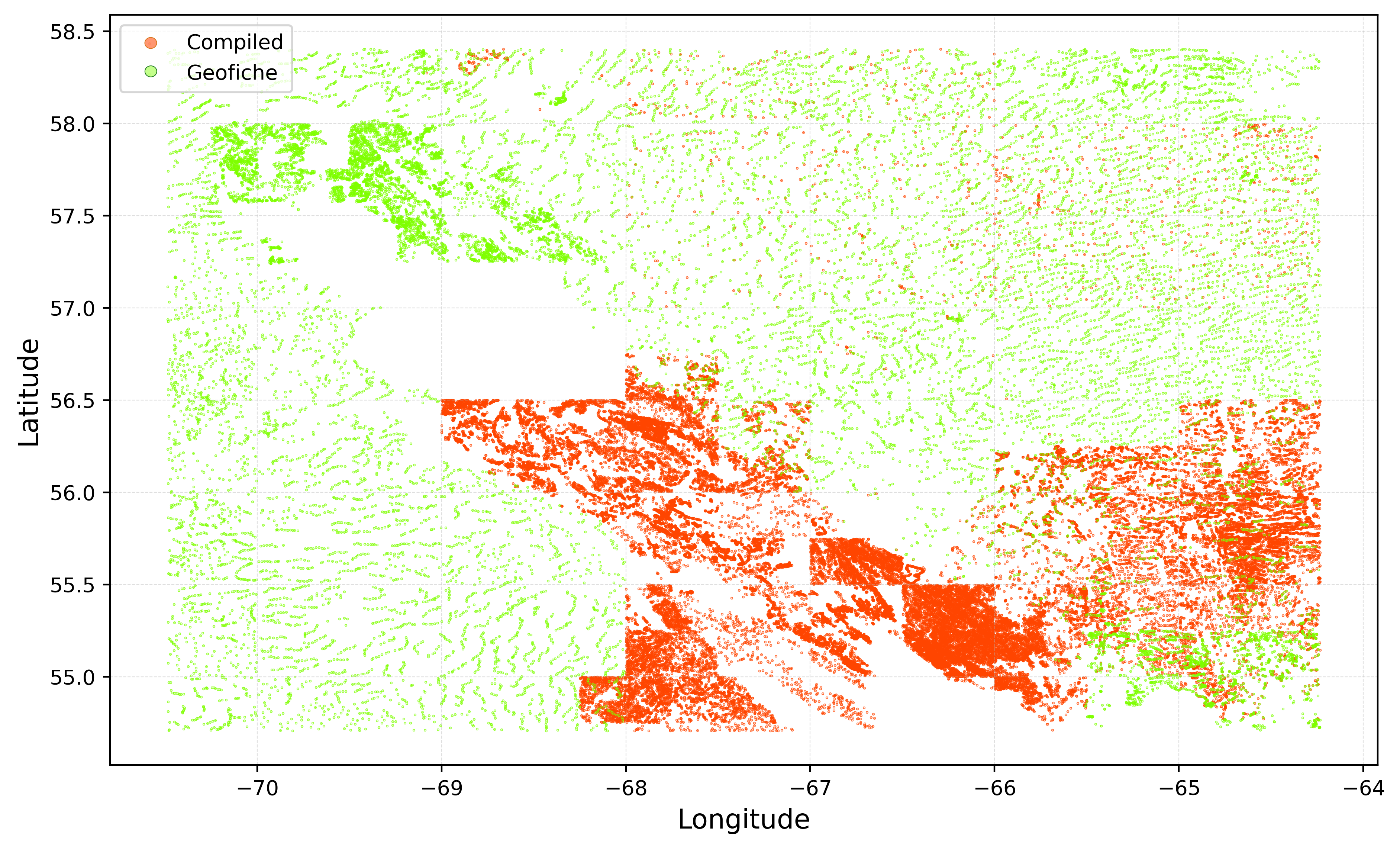}
        \caption{Datasets of rock observations in outcrops.}
        \label{subfig:outcrops_datasets}
    \end{subfigure}
    \hfill
    \begin{subfigure}[b]{0.70\linewidth}
        \centering
        \includegraphics[width=\linewidth]{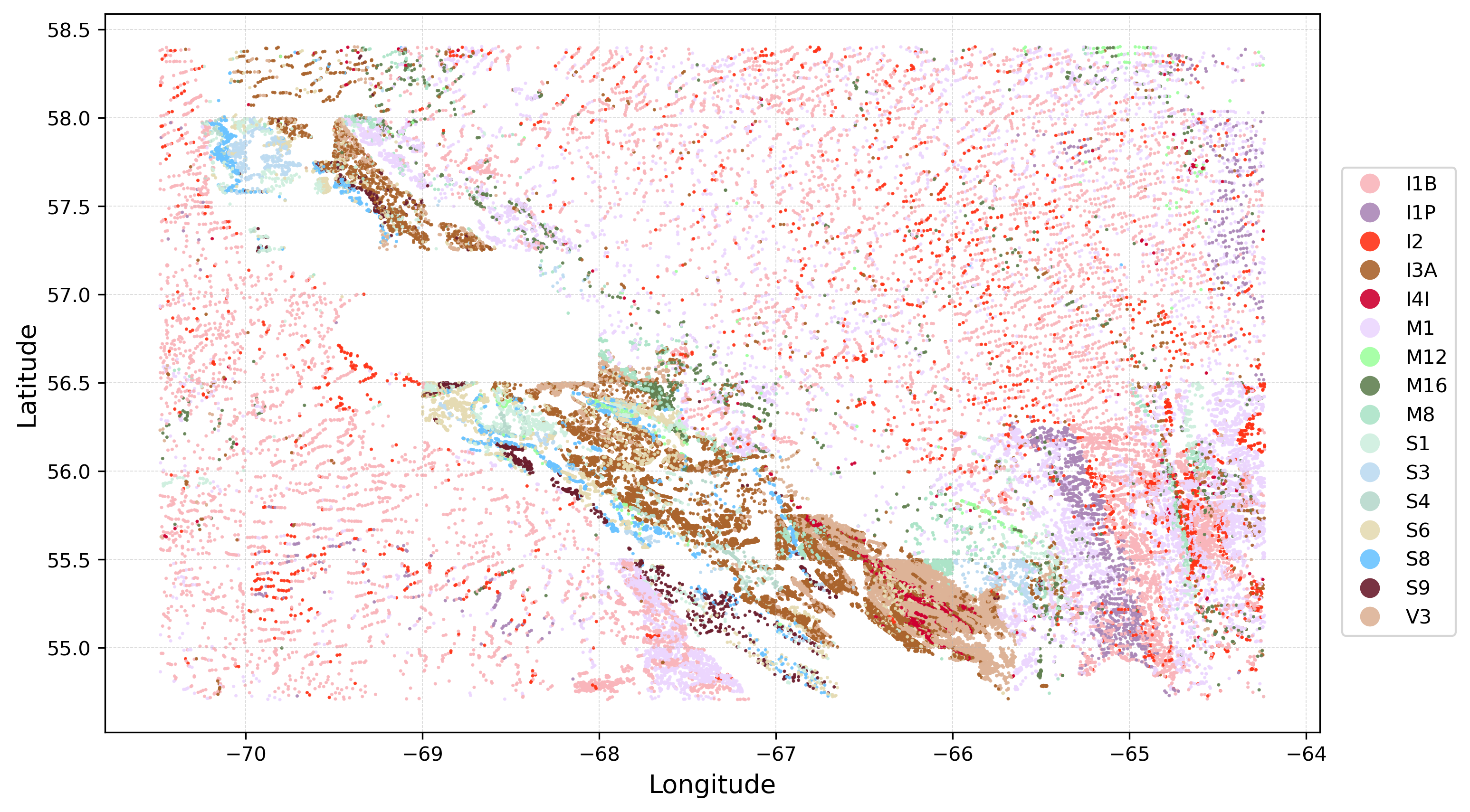}
        \caption{Lithologic units presenting frequency above 1\% in the northeast study area.}
        \label{subfig:field_samples}
    \end{subfigure}
    \caption{Field data available in the northeast study area.}
    \label{fig:ROCK_SAMPLES}
\end{figure}

The field data consists of point vectors that describe the lithologies observed in outcrops in Northern Quebec (Table \ref{tb:summary_samples_northeast}, Table \ref{tb:summary_samples_north}). We combined two datasets available in SIGÉOM: \textit{Affleurements de Geofiche (AG)} and \textit{Affleurements de Compilation} (AC). The AG dataset comprises field samples collected between 1967 and 2023, spaced at an average interval of 2.5 km. Meanwhile, the AC dataset compiles outcrops observed in the field and referenced in the literature, with an average sample spacing of 1 km. Additionally, only  samples with a frequency over 1\% were considered. The integrated dataset is visualized in Figure \ref{fig:ROCK_SAMPLES}. 

\begin{figure}
  \centering
  \includegraphics[width=0.8\textwidth]{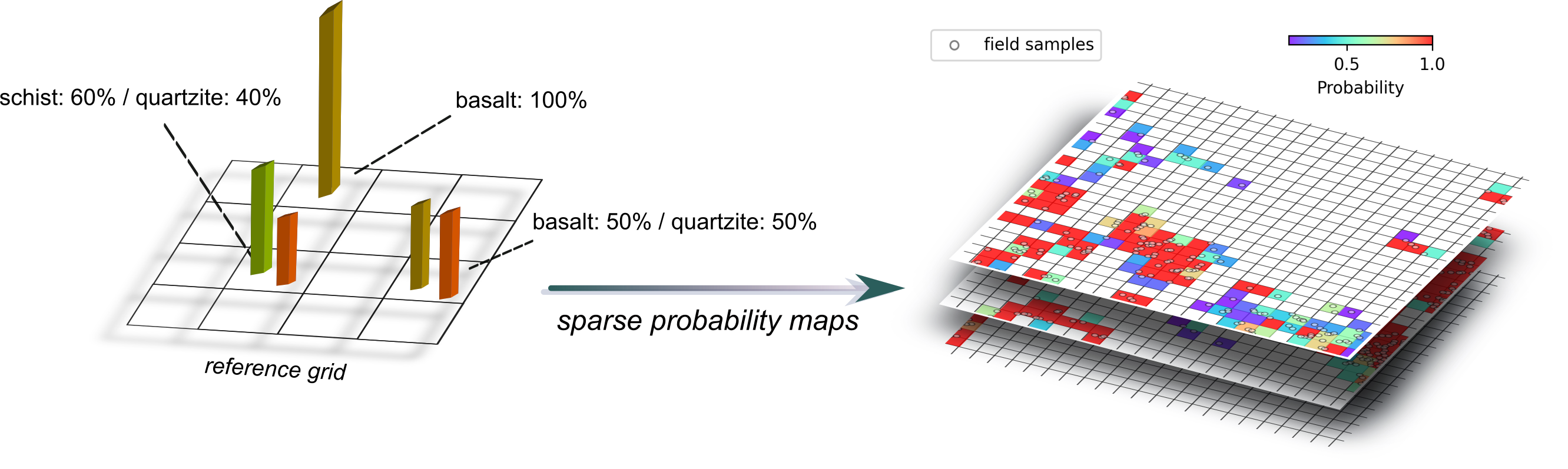}
  \caption{This figure illustrates the process of generating sparse probability masks from raw field data.}
  \label{fig:spr_probmasks}
\end{figure}

In order to use field data as targets and constraints for our model, we must map the samples onto the reference regular grid. For this purpose, we generated sparse probability masks through the calculation of class frequencies per pixel. In this approach, we basically populate a grid with probabilities between 0 and 1 depending on the frequencies of classes in a given pixel. That is, if a pixel contains only one class, it will receive probability 1.0. This process is better illustrated in Figure \ref{fig:spr_probmasks}.

\begin{table}
  \centering
  \caption{The codes, lithologies, and number of the samples observed in outcrops in the northeast area are displayed, with only lithologies having a frequency above 1\% shown.}\label{tb:summary_samples_northeast}
  \vspace{0.5cm} 
  \begin{tabular}{c|c|c}
    \hline
    \textbf{Code} & \textbf{Lithology} & \textbf{Number of Samples} \\
    \hline
    I1B & Granite & 11162\\
    I1P & Charnockite & 1700\\
    I2 & Intermediate Intrusive Rocks & 2439\\
    I3A & Gabbro & 13610\\
    I4I & Peridotite & 1422\\
    M1 & Gneiss & 12249\\
    M12 & Quartzite & 755\\
    M16 & Amphibolite & 1607\\
    M8 & Schist & 1607\\
    S1 & Sandstone & 2573\\
    S3 & Wacke & 1077\\
    S4 & Conglomerate & 610\\
    S6 & Mudrock & 3816\\
    S8 & Dolomite & 2141\\
    S9 & Iron Formation & 1183\\
    V3 & Mafic Volcanic Rocks & 9337\\
  \end{tabular}
\end{table}

\begin{table}
  \centering
  \caption{The codes, lithologies, and number of the samples observed in outcrops in the north area are displayed, with only lithologies having a frequency above 1\% shown.}\label{tb:summary_samples_north}
  \vspace{0.5cm} 
  \begin{tabular}{c|c|c}
    \hline
    \textbf{Code} & \textbf{Lithology} & \textbf{Number of Samples} \\
    \hline
    I1 & Intrusive Felsic Rocks & 2491\\
    I2 & Intermediate Intrusive Rocks & 610\\
    I3 & Intrusive Mafic Rocks & 2932\\
    I4 & Intrusive Ultramafic Rocks & 422\\
    M1 & Gneiss & 776\\
    M8 & Schist & 1248\\
    V3 & Mafic Volcanic Rocks & 4203\\
  \end{tabular}
\end{table}

\section{Results}

In this section, we present the results of our approach within selected areas in Northern Quebec. It includes predictive lithological maps,  evaluation of the model's performance in both training and validation datasets, and prediction uncertainty evaluation. 

\subsection{Northeast study area}
\begin{figure}
  \centering
  \includegraphics[width=0.75\textwidth]{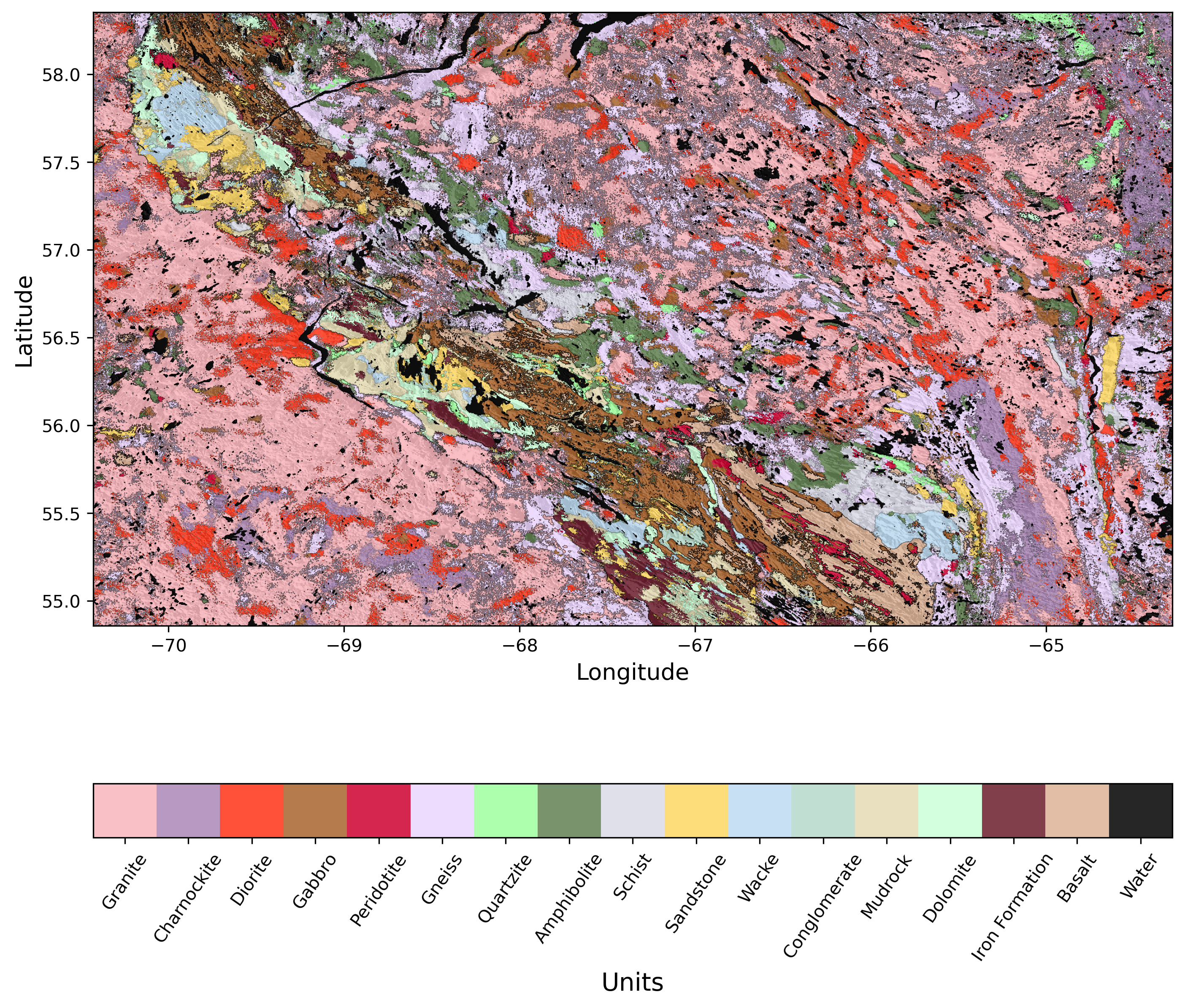}
  \caption{Predictive lithological map of the northeast area presenting 16 lithologic units. This map represents the average of 100 spatially constrained predictions.}
  \label{fig:litomap_northeast}
\end{figure}

\begin{figure}
  \centering
  \includegraphics[width=0.8\textwidth]{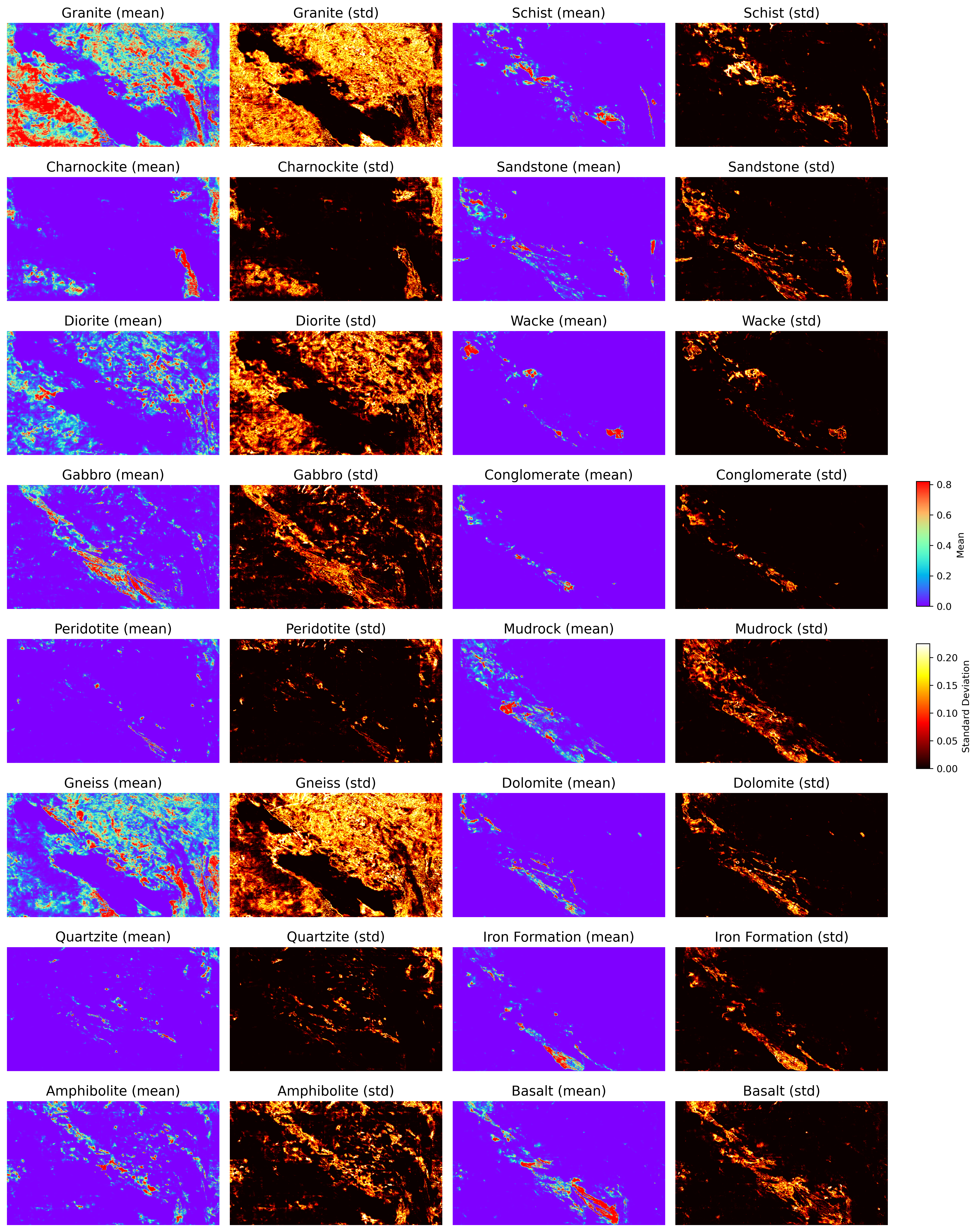}
  \caption{Mean and standard deviation of the predictions for each lithologic unit.}
  \label{fig:mean_std_map}
\end{figure}

\begin{figure}
  \centering
  \begin{subfigure}{0.5\linewidth}
    \includegraphics[width=\linewidth]{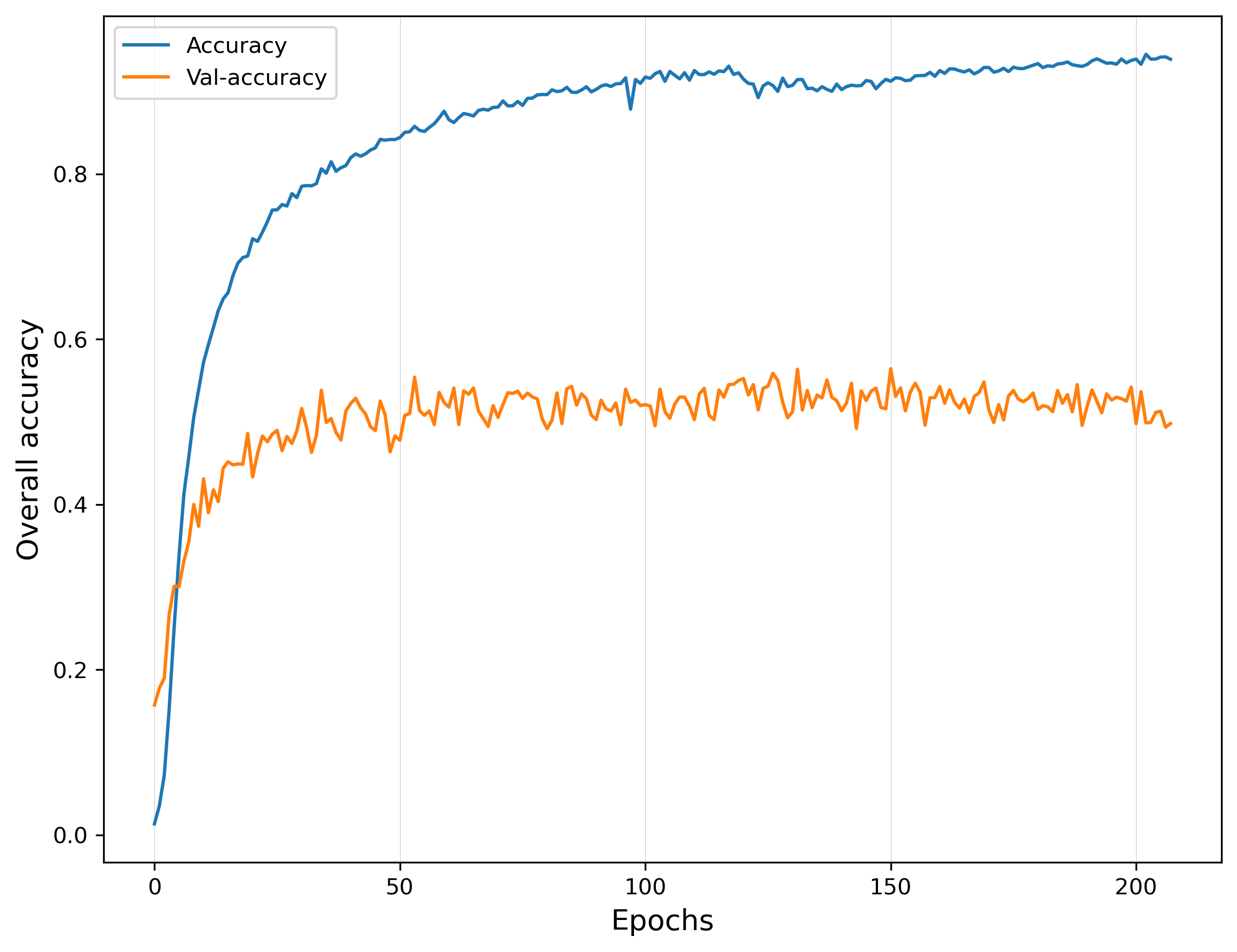}
    \caption{Overall accuracy of training and testing sets.}
  \end{subfigure}%
  \hfill
  \begin{subfigure}{0.5\linewidth}
    \includegraphics[width=\linewidth]{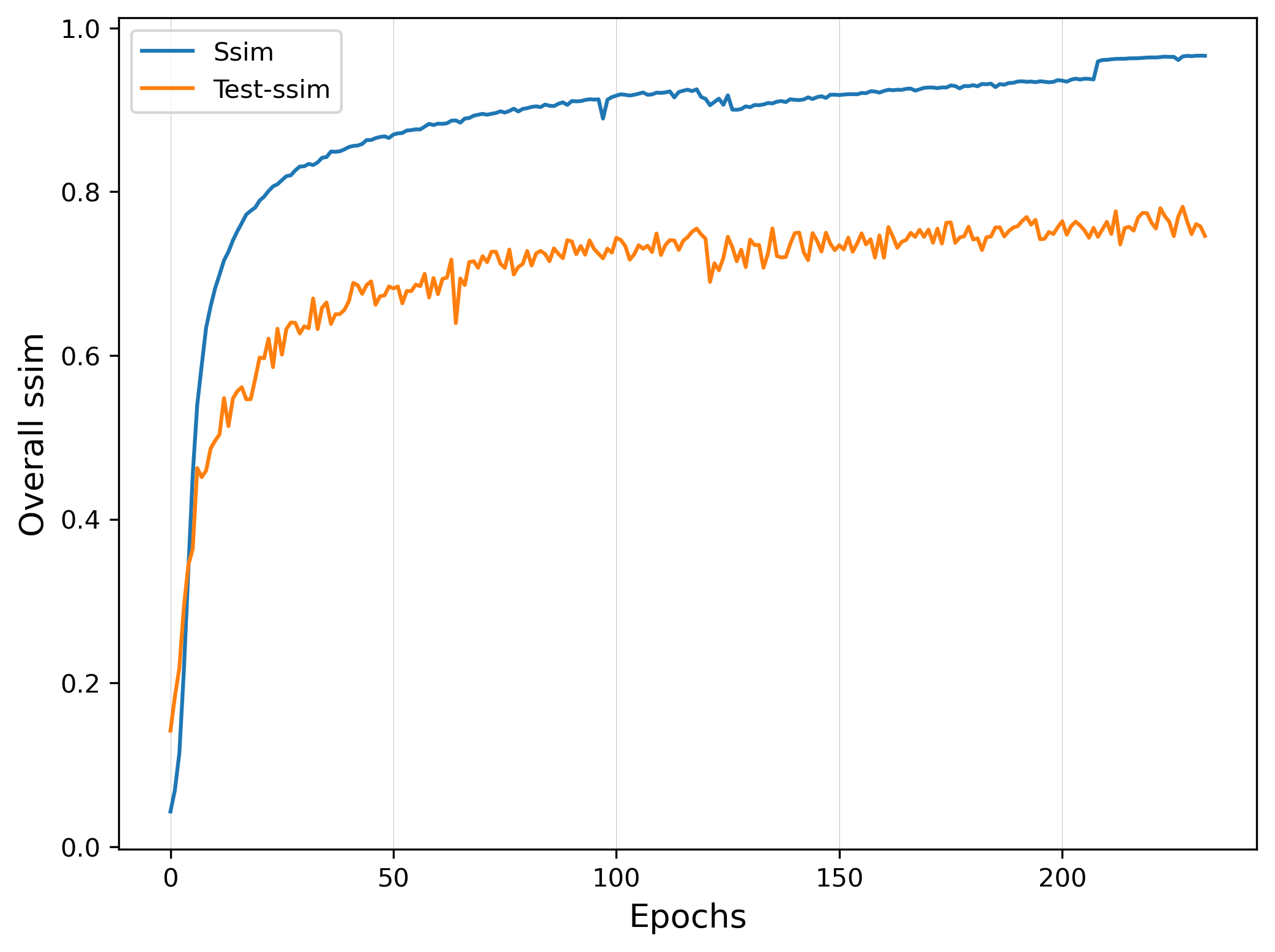}
    \caption{Overall structural similarity of training and testing sets.}
  \end{subfigure}
  \caption{The graphs depict accuracy and structural similarities in training and testing datasets (Northeast area).}
  \label{fig:fig_metrics}
\end{figure}

\begin{figure}
  \centering
  \begin{subfigure}{0.5\linewidth}
    \includegraphics[width=\linewidth]{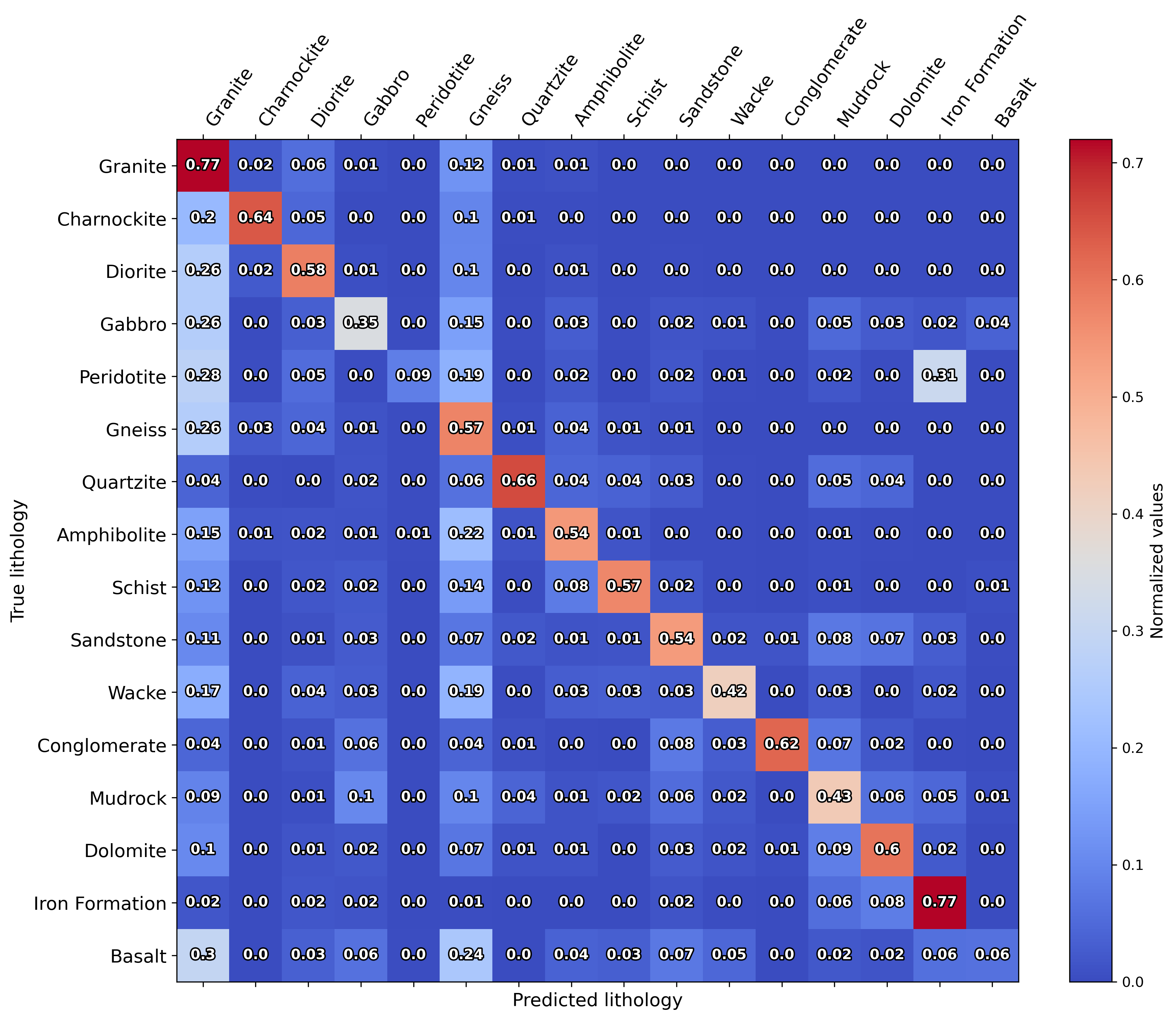}
    \caption{Training set: unconstrained predictions.}
  \end{subfigure}%
  \hfill
  \begin{subfigure}{0.5\linewidth}
    \includegraphics[width=\linewidth]{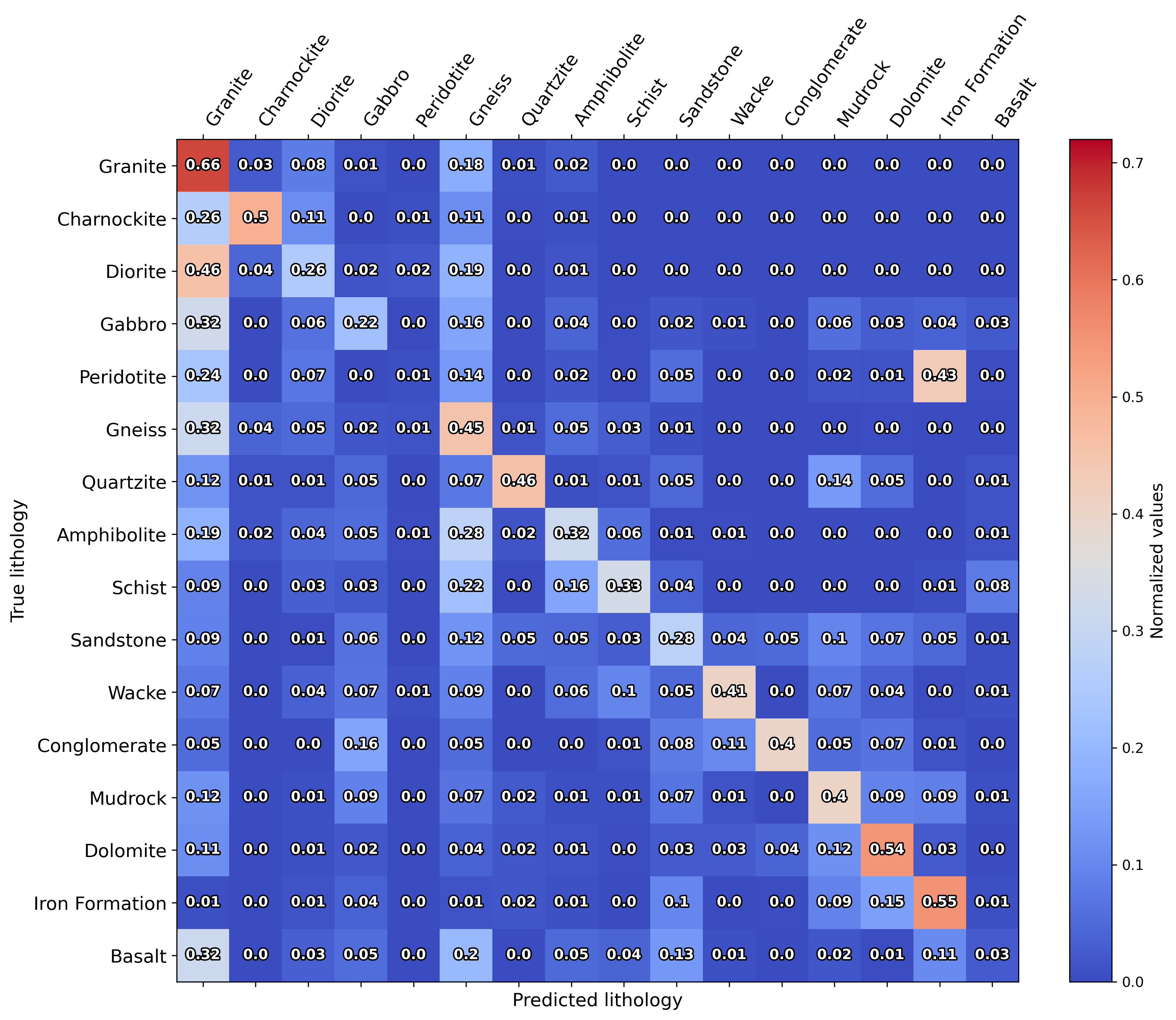}
    \caption{Testing set: unconstrained predictions.}
  \end{subfigure}
  \\
  \begin{subfigure}{0.5\linewidth}
    \includegraphics[width=\linewidth]{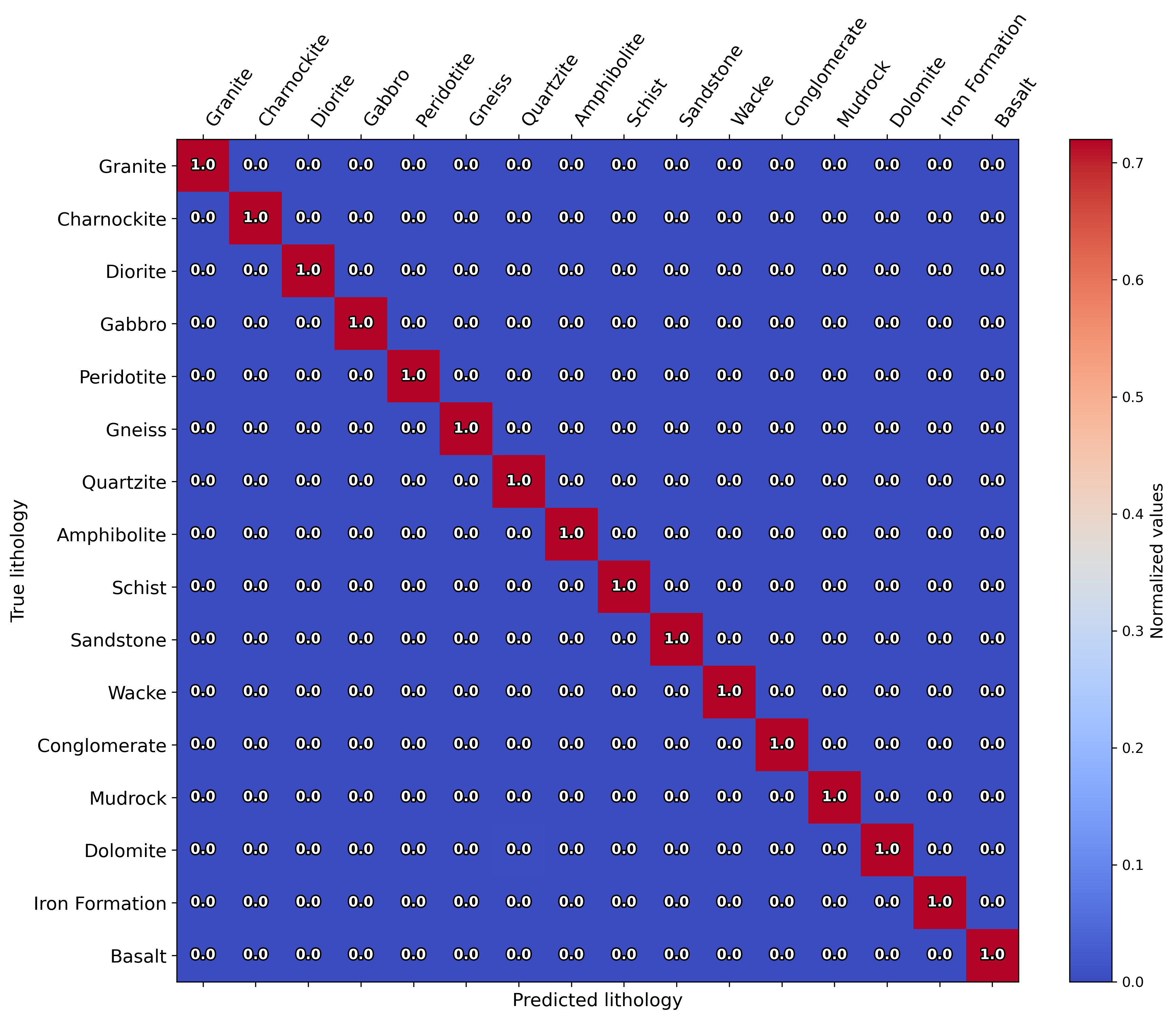}
    \caption{Training set: constrained predictions.}
  \end{subfigure}%
  \hfill
  \begin{subfigure}{0.5\linewidth}
    \includegraphics[width=\linewidth]{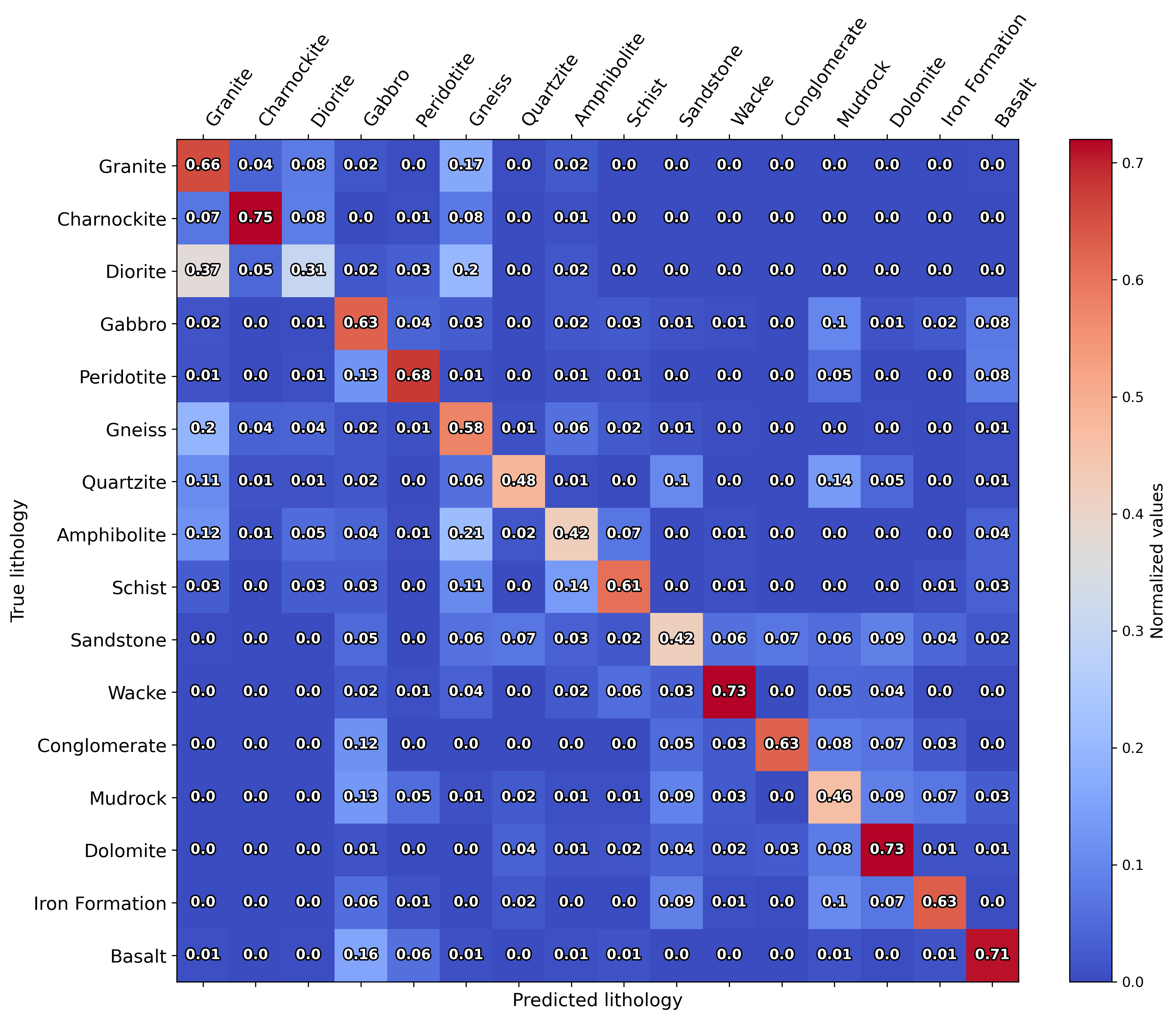}
    \caption{Testing set: constrained predictions.}
  \end{subfigure}
  \caption{Confusion matrices for the training and testing datasets in the northeast area. The first row shows the results for unconstrained predictions, while the second row shows the results for predictions constrained by training samples.}
  \label{fig:cfm_northeast}
\end{figure}

\begin{figure}
  \centering
  \begin{subfigure}{0.7\linewidth}
    \includegraphics[width=\linewidth]{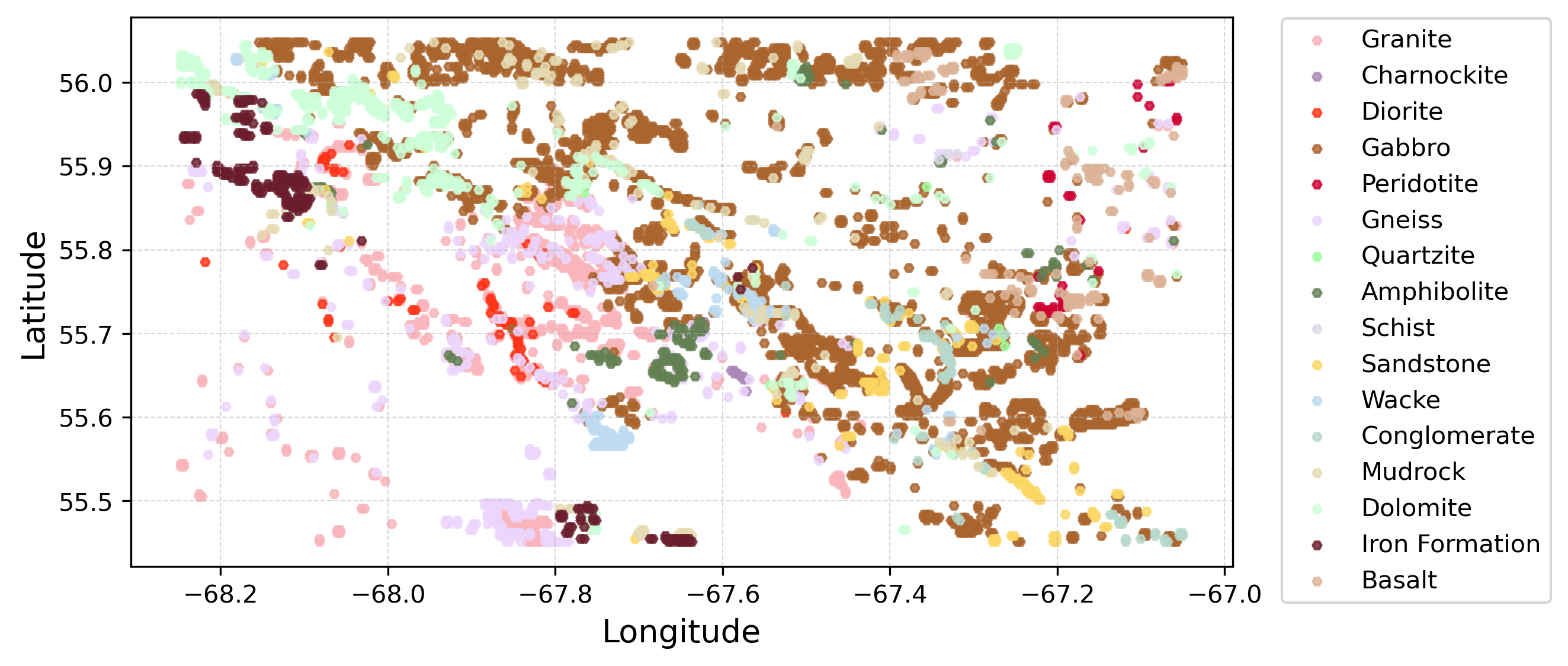}
    \caption{Predictions.}
  \end{subfigure}
  \hfill
  \begin{subfigure}{0.7\linewidth}
    \includegraphics[width=\linewidth]{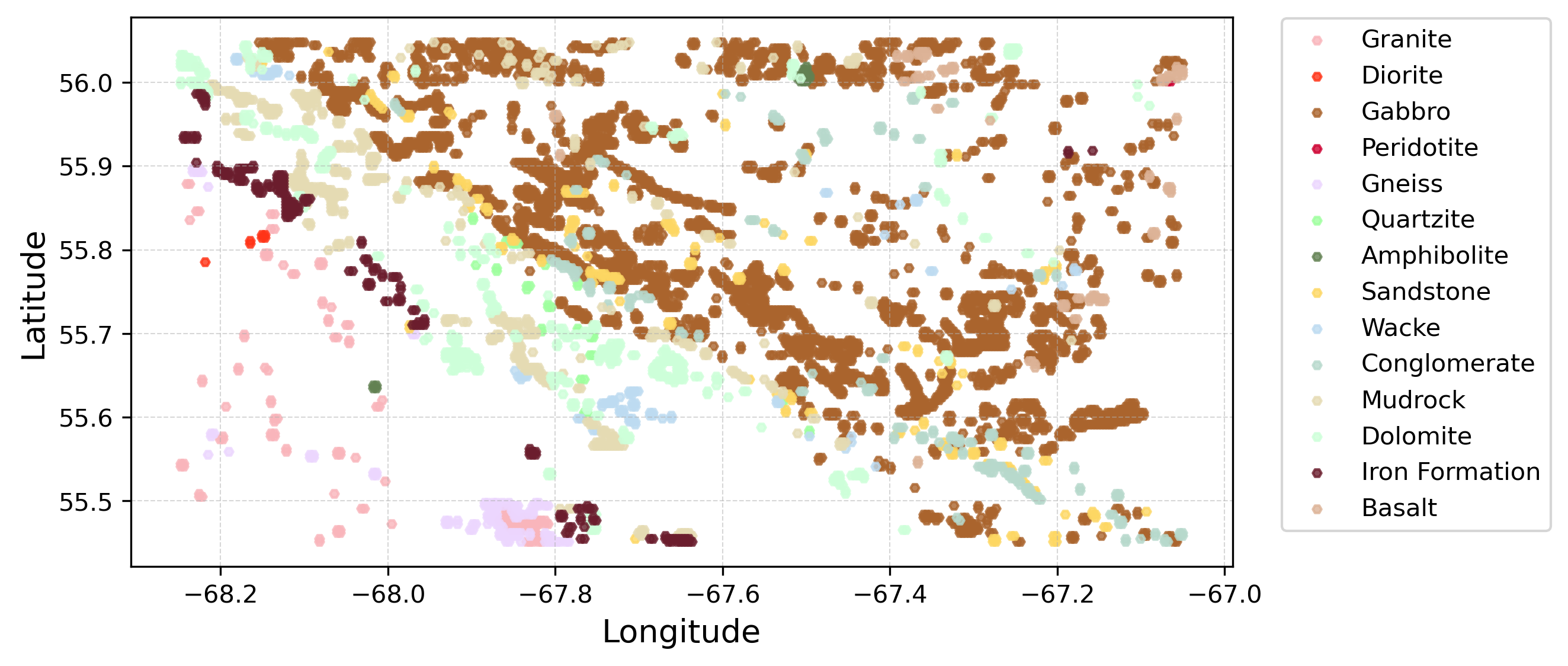}
    \caption{Ground truth.}
  \end{subfigure}%
  \hfill
  \begin{subfigure}{0.7\linewidth}
    \includegraphics[width=\linewidth]{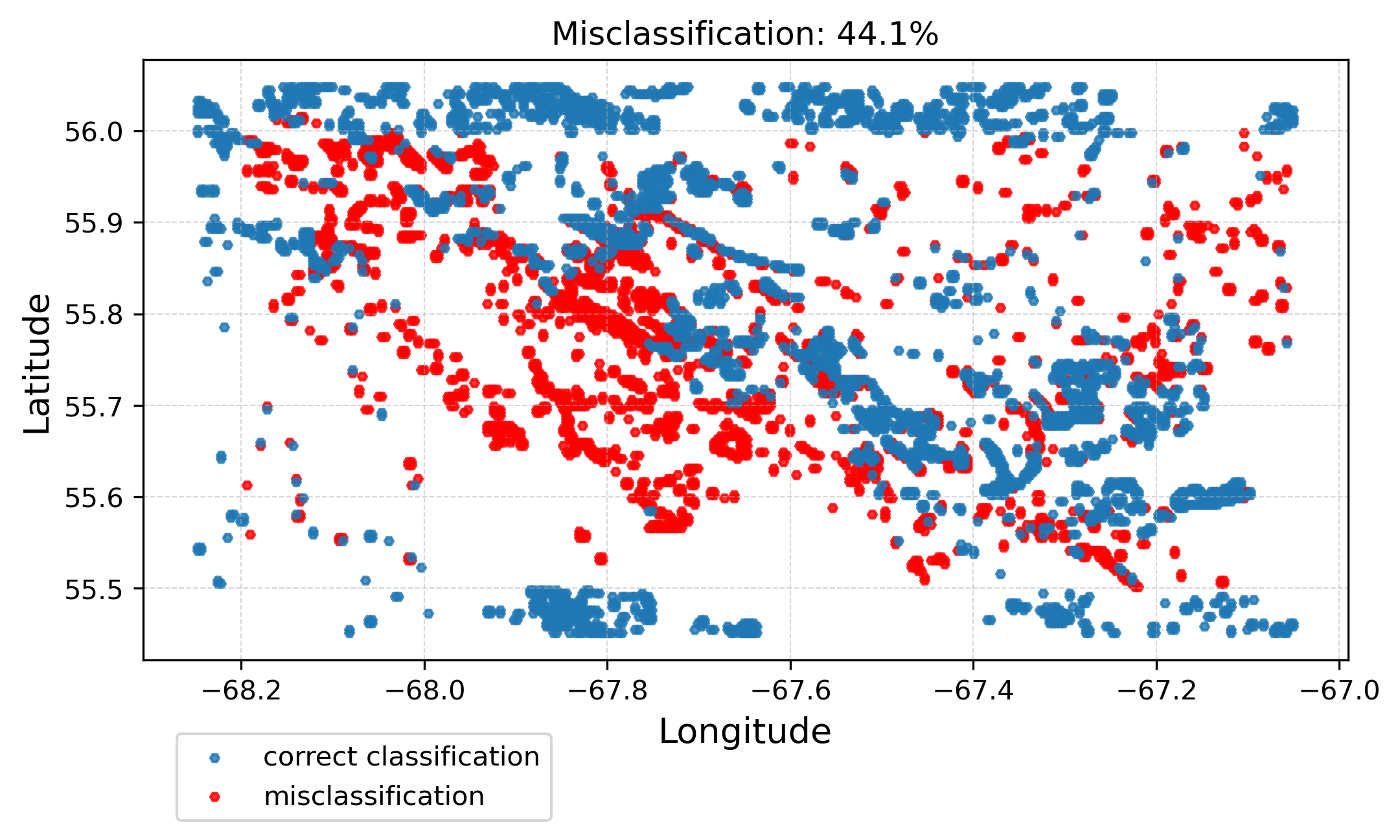}
    \caption{Misclassifications.}
  \end{subfigure}%
  \caption{The figures show, from top to bottom, misclassified samples in red and correctly classified samples in blue, along with predicted and ground truth classes in the validation area.}
  \label{fig:val_area}
\end{figure}

Figure \ref{fig:litomap_northeast} illustrates the average predictive lithological map for the northeast region. This map resulted from averaging 100 predictive maps generated using the Monte Carlo Dropout technique. In Figure \ref{fig:mean_std_map}, we depict the mean and standard deviation derived from all 100 predictions per class. Therefore, the lithologies displayed in Figure \ref{fig:litomap_northeast} represent the most probable predicted lithologic units on average. Conversely, the standard deviation maps indicate locations where the most probable class varies among predictions,

To evaluate the overall performance, Figure \ref{fig:fig_metrics} displays overall weighted accuracy and structural similarity for both training and testing datasets. It shows that the model converges after about 150 epochs, achieving an overall weighted accuracy of over 0.9 and 0.5 for the training and validation sets, respectively. In order to check the performance of the model on classifying each of the rock types, confusion matrices in Figure \ref{fig:cfm_northeast} show the principal diagonals representing the normalized rate of correctly predicted samples. The matrices present a comparison between predictions obtained with and without spatial constraints. In Figure \ref{fig:val_area}, correct and incorrect classifications are shown in the validation area along with the ground truth and predictions for comparison. This serves to analyze the model's performance in a spatially uncorrected zone of the northeast area.

\subsection{North study area (Transfer learning strategy)}

\begin{figure}
  \centering
  \includegraphics[width=0.75\textwidth]{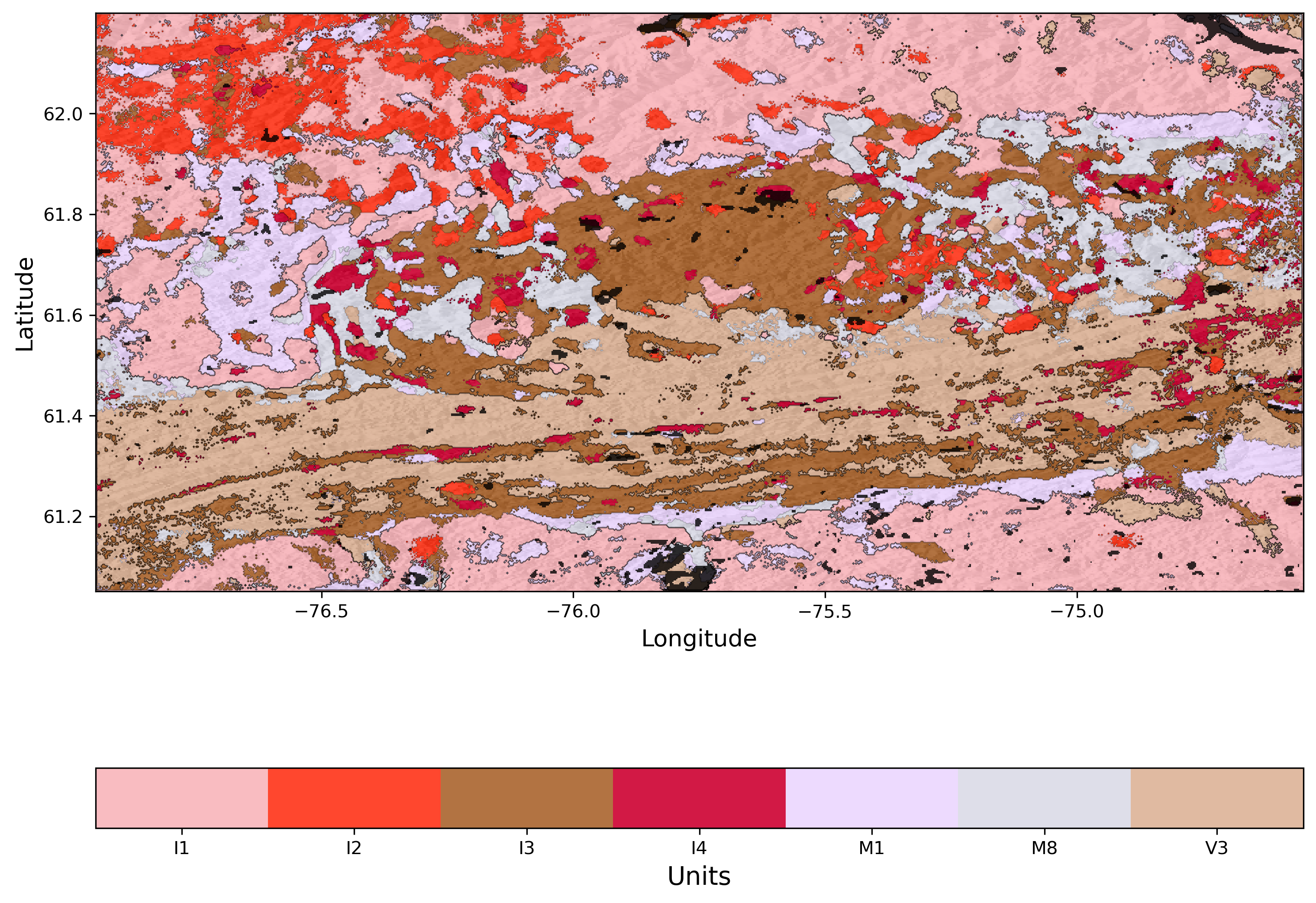}
  \caption{Predictive lithological map of the north area. The map represents 7 distinct lithologic units and was generated through fine-tuning the model pre-trained on the northeast area dataset.}
  \label{fig:litomap_north}
\end{figure}

\begin{figure}
  \centering
  \begin{subfigure}{0.45\linewidth}
    \includegraphics[width=\linewidth]{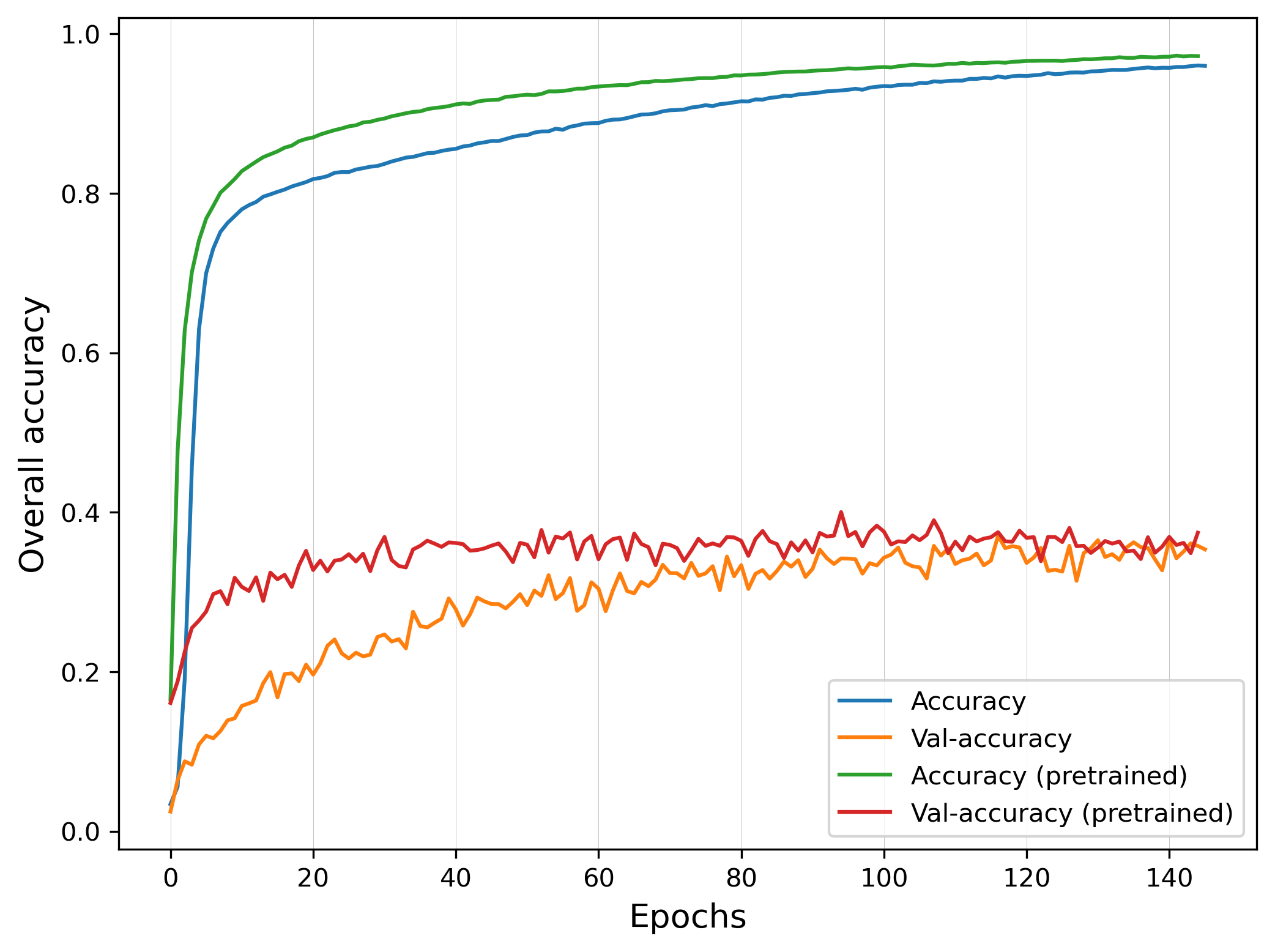}
    \caption{Performance comparison: Overall accuracy.}
  \end{subfigure}%
  \hfill
  \begin{subfigure}{0.45\linewidth}
    \includegraphics[width=\linewidth]{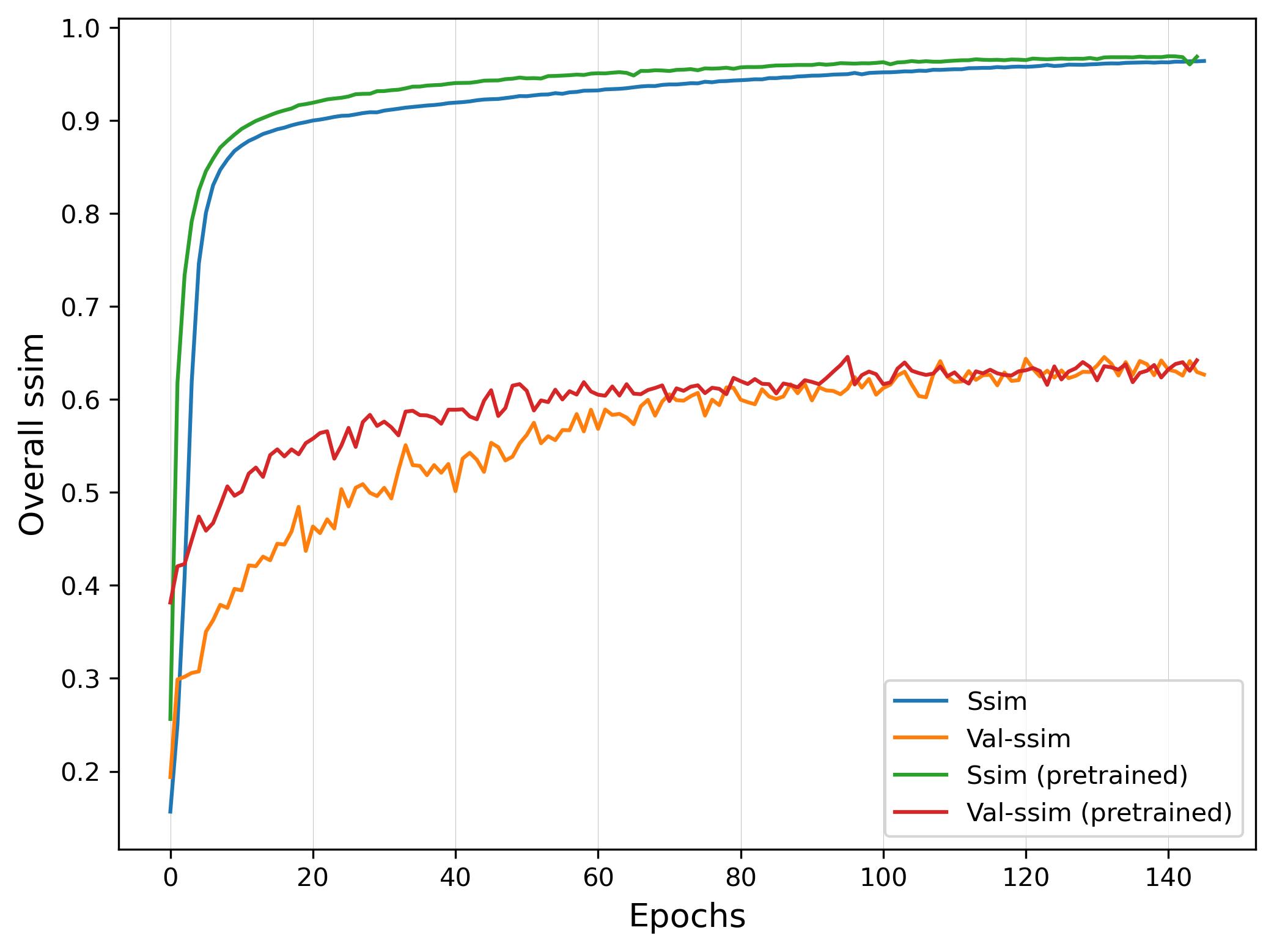}
    \caption{Performance comparison: Overall structural similarity.}
  \end{subfigure}
  \caption{The graphs compare the performances obtained with a model trained from scratch (blue and orange curves) and another with pretrained weights from the northeast area (green and red curves).}
  \label{fig:tl_comparison}
\end{figure}

The image in Figure \ref{fig:litomap_north} shows the predictive lithological map for the northern area, which was generated using a transfer learning approach. To create this map, we averaged 100 predictive maps that were generated using different random dropout configurations. We fine-tuned a pre-trained model, which was originally trained on the northeast area, using the dataset from the northern area. The lithologies shown in Figure \ref{fig:litomap_north} represent the most probable predicted lithologic units, on average. As shown in Figure \ref{fig:tl_comparison}, the model was able to converge after only 50 epochs of fine-tuning, which is a significant improvement compared to the 90 epochs that were required to train the model from scratch. Additionally, using pre-trained weights resulted in an overall performance increase of 3\%.

\begin{figure}
  \centering
  \begin{subfigure}{0.5\linewidth}
    \centering
    \includegraphics[width=\linewidth]{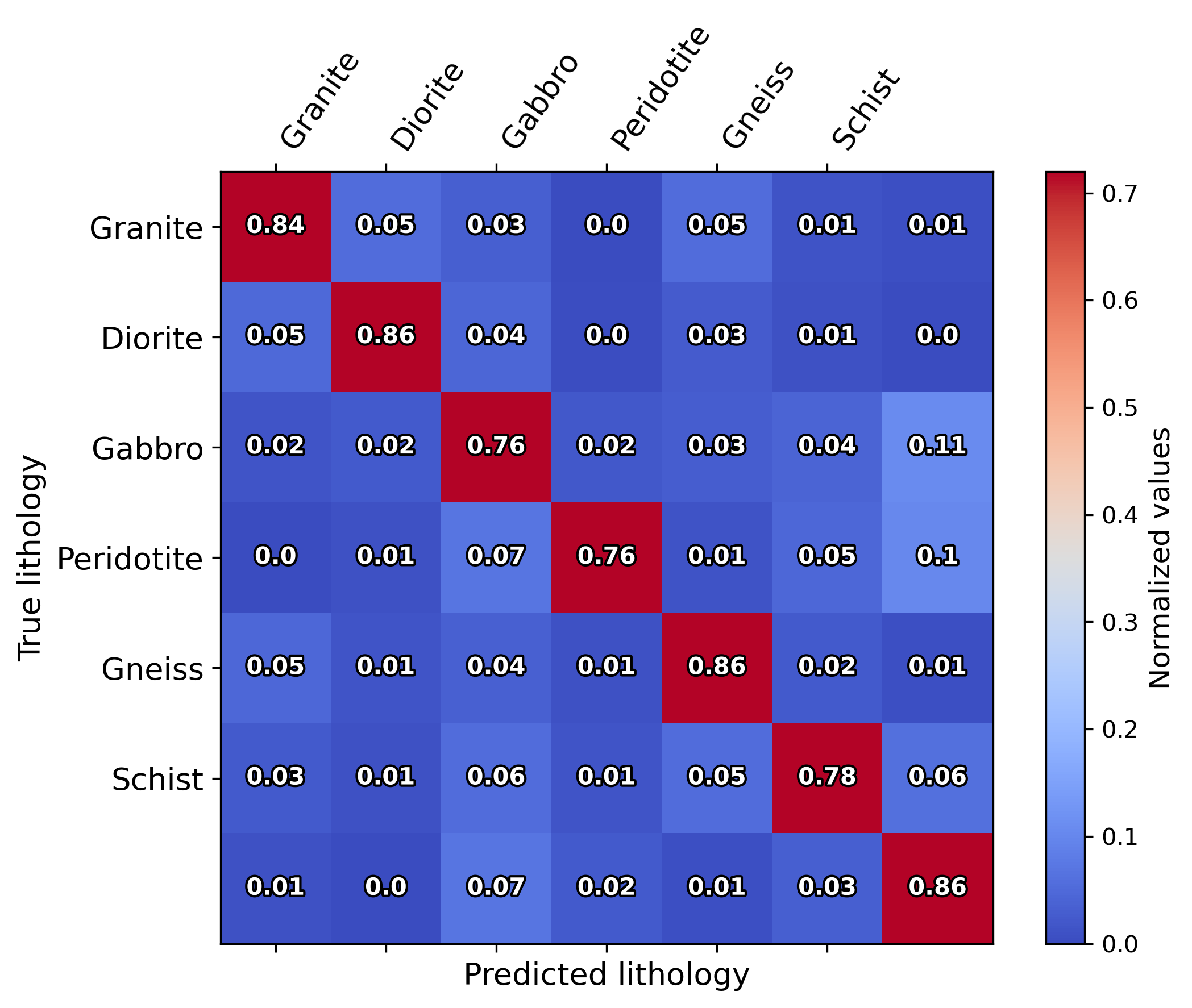}
    \caption{Training set: unconstrained predictions.}
  \end{subfigure}%
  \hfill
  \begin{subfigure}{0.5\linewidth}
    \centering
    \includegraphics[width=\linewidth]{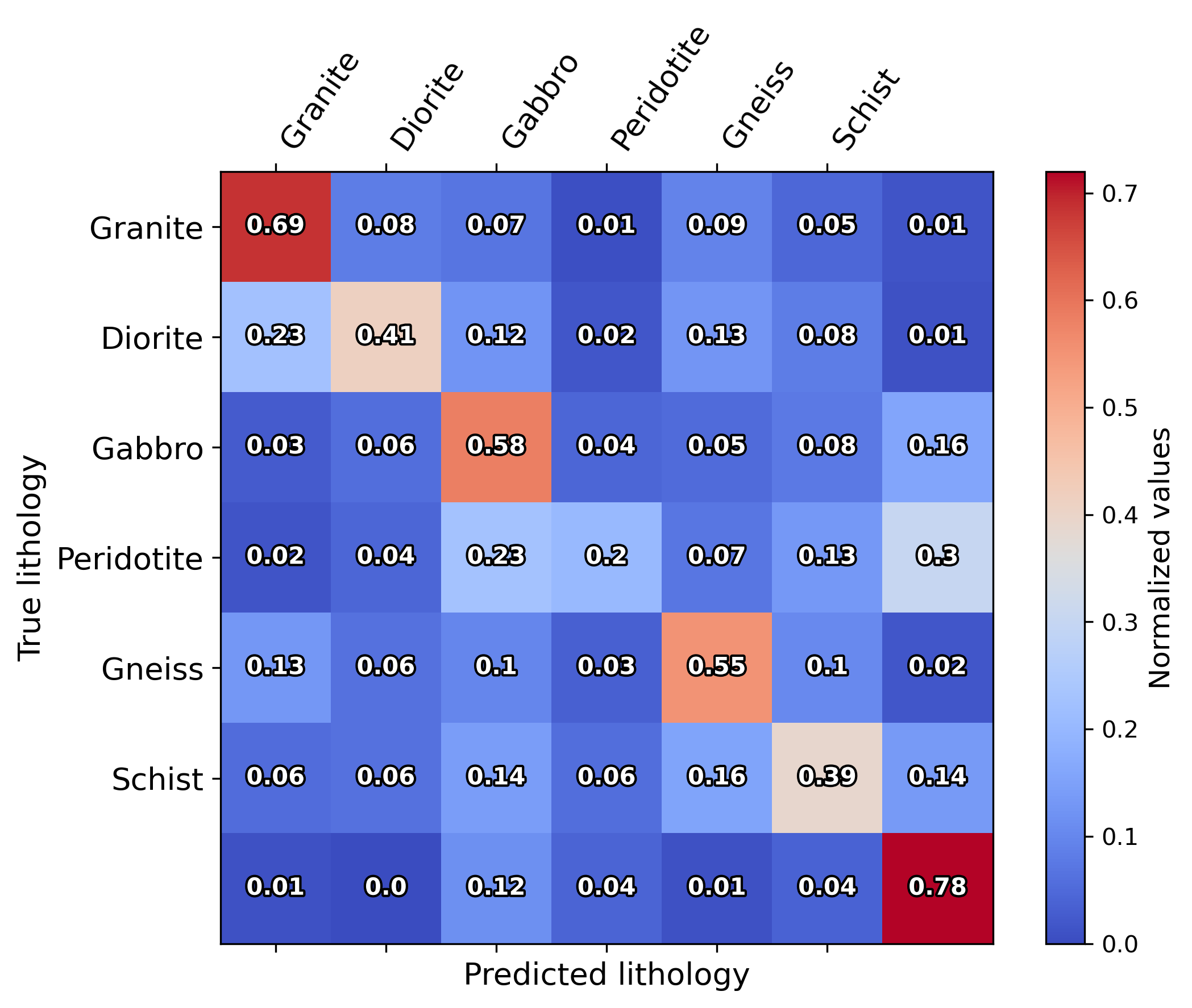}
    \caption{Validation set: unconstrained predictions.}
  \end{subfigure}
  \\
  \begin{subfigure}{0.5\linewidth}
    \centering
    \includegraphics[width=\linewidth]{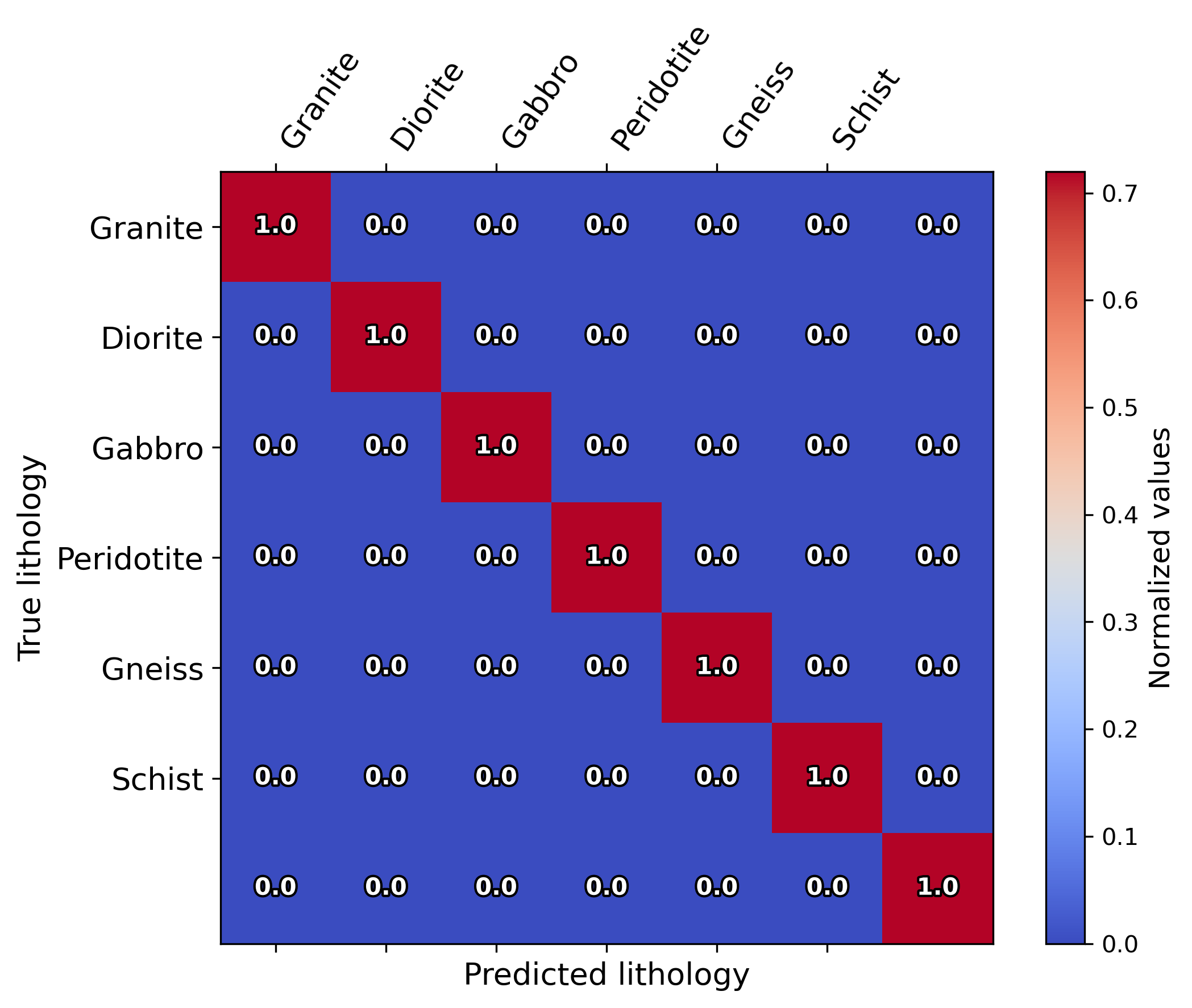}
    \caption{Training set: constrained predictions.}
  \end{subfigure}%
  \hfill
  \begin{subfigure}{0.5\linewidth}
    \centering
    \includegraphics[width=\linewidth]{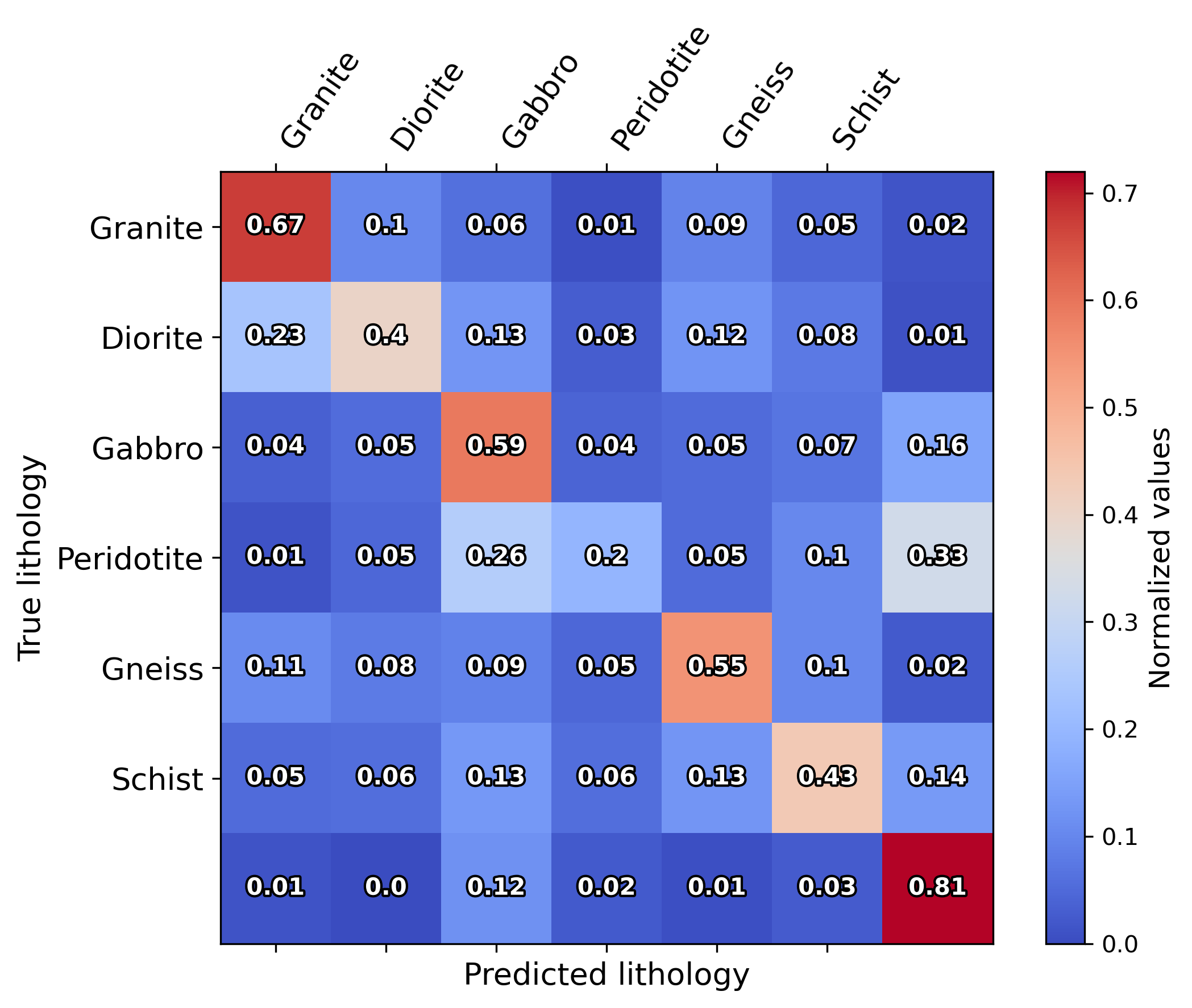}
    \caption{Validation set: constrained predictions.}
  \end{subfigure}
  \caption{Above are the confusion matrices for the training and validation datasets in the north area. The first row shows the results for unconstrained predictions, while the second row shows the results for predictions constrained by training samples.}
  \label{fig:cfm_north}
\end{figure}

In Figure \ref{fig:cfm_north}, confusion matrices present a comparison between lithology predictions constrained and unconstrained by training probability masks.

\section{Discussion}

In the northeast area, SCB-Net predicted 16 different types of lithology units. When predictions were constrained by sparse probability masks, ten of these units showed an overall weighted accuracy of over 0.6 in the testing set. This was a significant improvement compared to just one unit with unconstrained predictions. The use of probability masks improved the accuracy of predictions for most rock types in the northeast area, such as charnockite, gabbro, peridotite, dolomite, wacke, and basalt. It is interesting to note that rocks presenting very similar compositions (predominantly quartz and feldspar minerals), such as granite, schist, gneiss, diorite, and sandstone, were greatly impacted by the use of spatial constraints. This demonstrates the importance of spatially explicit information on prediction accuracy, particularly when rock types have very similar physical properties and cannot be clearly distinguished using auxiliary variables. In the northern area, some units were not noticeably affected by spatial limitations. For instance, Granite experienced a 2\% decline in its performance when such constraints were applied. This indicates that spatially explicit data can sometimes force spatial consistency where it does not exist, which underscores the importance of being cautious when selecting filter sizes and weights for the dilation operator. Another drawback of using this operator is that it is challenging to establish a unique set of weights to optimize all classes, meaning that we must rely  on the model's learned filters to generate predictions with varying spatial consistencies by class.

During the training process, sparse probability masks are utilized to constrain lithology predictions based on field samples. In the training set, all samples are predicted with 100\% accuracy when using these masks. However, this is not the case when zeroed probability masks are utilized. This demonstrates the model's ability to learn spatial patterns from the ground-truth data while avoiding over-fitting to the training dataset. In essence, this approach has the capacity to generate predictions that are constrained by field data while still being able to generalize to unseen data.

In the validation set of the northeast area, the model struggled to predict sedimentary rocks, such as dolomite, mudrock, conglomerate, and wacke, while exhibiting satisfactory results for intrusive and volcanic mafic rocks, as well as granite. This reinforces the notion that without spatially explicit information, rock types with low contrast in physical properties are often misclassified.

Block dropout was only applied to the first part of our network, which is responsible for extracting features from the auxiliary data. This strategy effectively reduces prediction variance at sampled locations, thereby directing uncertainty quantification towards locations with no samples. As a result,  isolated dots may appear in certain areas, and it's up to experts to determine whether they represent noise or highly discontinuous rock formations. This is particularly important in the context of predictive mapping, where decision-making regarding areas with none or few samples must be made.

The use of transfer learning strategy in the northern area has shown that it has the potential to generate predictions for previously unobserved areas after fine-tuning. The model was able to converge after only 50 epochs when using pre-trained model weights, compared to 90 epochs when trained from scratch. This approach has led to a 50\% reduction in training time, which is a significant decrease in computational cost. Additionally, there was an overall performance improvement of 3\%, which demonstrates the effectiveness of SCB-Net in transfer learning, at least in areas with similar geological contexts.

\section{Conclusions}

In this study, we introduced a novel approach for lithological mapping that combines the strengths of classical predictive mapping methods (e.g., geostatistical techniques like indicator cokriging), enabling the generation of field-data-constrained predictions and  uncertainty quantification, while benefiting from the capabilities of machine learning methods in multivariate problems. Additionally, our approach harnesses the power of deep learning methods to extract complex non-linear relationships present in the data. This approach is particularly useful in scenarios involving the presence of historical field data and auxiliary information, such as satellite imagery and airborne geophysical data, which are often the case in regional mapping and exploration campaigns.

The SCB-Net's capability was demonstrated to estimate the spatial distribution of 16 distinct lithologic units in the northeast area and 7 lithologic units in the north area using both explicit spatial information and remotely sensed data. The model's potential to generate predictions for previously unobserved areas was also tested using a transfer learning strategy. The results indicate that the pre-trained SCB-Net model from the northeast area can make accurate predictions in the northern area after fine-tuning, while reducing the computational cost of model training.
 
Future work should prioritize the application of SCB-Net to diverse geological contexts and large-scale mapping, where field data is typically more limited. This should also involve the integration of additional auxiliary information, such as geochemical data. Furthermore, the use of transfer learning in different geological contexts and datasets should be explored further.

\section{Acknowledgments}

We thank the Geological Survey of Canada's GEM-GeoNorth program, a part of Natural Resources Canada (NRCan), for providing full funding for this project.

\newpage

\textbf{Code availability section}

SCB-Net

Contact: victor.santos@inrs.ca or victor.silva.santos@alumni.usp.br

Hardware requirements: GPU with at least 8 GB of memory

Program language: Python
 
Library required: Tensorflow, Numpy, Matplotlib, Scikit-learn, Geopandas, Rasterio, and others

The source codes are available for downloading at the link:
https://github.com/victsnet/SCB-Net

\bibliographystyle{plainnat}  

\bibliography{bibliography}

\begin{thebibliography}{24}
\providecommand{\natexlab}[1]{#1}
\providecommand{\url}[1]{\texttt{#1}}
\expandafter\ifx\csname urlstyle\endcsname\relax
  \providecommand{\doi}[1]{doi: #1}\else
  \providecommand{\doi}{doi: \begingroup \urlstyle{rm}\Url}\fi

\bibitem[Abadi(2016)]{abadi2016tensorflow}
Mart{\'\i}n Abadi.
\newblock Tensorflow: learning functions at scale.
\newblock In \emph{Proceedings of the 21st ACM SIGPLAN international conference on functional programming}, pages 1--1, 2016.

\bibitem[Cedou et~al.(2022)Cedou, Gloaguen, Blouin, Cat{\'e}, Paiement, and Tirdad]{cedou2022preliminary}
Matthieu Cedou, Erwan Gloaguen, Martin Blouin, Antoine Cat{\'e}, Jean-Philippe Paiement, and Shiva Tirdad.
\newblock Preliminary geological mapping with convolution neural network using statistical data augmentation on a 3d model.
\newblock \emph{Computers \& Geosciences}, 167:\penalty0 105187, 2022.

\bibitem[Costa et~al.(2019)Costa, Tavares, and de~Oliveira]{costa2019predictive}
Iago Sousa~Lima Costa, Felipe~Mattos Tavares, and Junny Kyley~Mastop de~Oliveira.
\newblock Predictive lithological mapping through machine learning methods: a case study in the cinzento lineament, caraj{\'a}s province, brazil.
\newblock \emph{Journal of the Geological Survey of Brazil}, 2\penalty0 (1):\penalty0 26--36, 2019.

\bibitem[Cracknell and Reading(2014)]{cracknell2014geological}
Matthew~J Cracknell and Anya~M Reading.
\newblock Geological mapping using remote sensing data: A comparison of five machine learning algorithms, their response to variations in the spatial distribution of training data and the use of explicit spatial information.
\newblock \emph{Computers \& Geosciences}, 63:\penalty0 22--33, 2014.

\bibitem[Cuba et~al.(2009)Cuba, Babak, and Leuangthong]{cuba2009selection}
Miguel Cuba, Olena Babak, and Oy~Leuangthong.
\newblock On the selection of secondary variables for cokriging and cosimulation.
\newblock \emph{CCG Annual Report}, 11, 2009.

\bibitem[Dietterich(1995)]{dietterich1995overfitting}
Tom Dietterich.
\newblock Overfitting and undercomputing in machine learning.
\newblock \emph{ACM computing surveys (CSUR)}, 27\penalty0 (3):\penalty0 326--327, 1995.

\bibitem[Gal and Ghahramani(2016)]{gal2016dropout}
Yarin Gal and Zoubin Ghahramani.
\newblock Dropout as a bayesian approximation: Representing model uncertainty in deep learning.
\newblock In \emph{international conference on machine learning}, pages 1050--1059. PMLR, 2016.

\bibitem[Ghahramani(2015)]{ghahramani2015probabilistic}
Zoubin Ghahramani.
\newblock Probabilistic machine learning and artificial intelligence.
\newblock \emph{Nature}, 521\penalty0 (7553):\penalty0 452--459, 2015.

\bibitem[Ghiasi et~al.(2018)Ghiasi, Lin, and Le]{ghiasi2018dropblock}
Golnaz Ghiasi, Tsung-Yi Lin, and Quoc~V Le.
\newblock Dropblock: A regularization method for convolutional networks.
\newblock \emph{Advances in neural information processing systems}, 31, 2018.

\bibitem[Goyal(2011)]{goyal2011morphological}
Megha Goyal.
\newblock Morphological image processing.
\newblock \emph{IJCST}, 2\penalty0 (4):\penalty0 59, 2011.

\bibitem[Guart{\'a}n and Emery(2021)]{guartan2021predictive}
Jos{\'e}~A Guart{\'a}n and Xavier Emery.
\newblock Predictive lithological mapping based on geostatistical joint modeling of lithology and geochemical element concentrations.
\newblock \emph{Journal of Geochemical Exploration}, 227:\penalty0 106810, 2021.

\bibitem[Harris and Grunsky(2015)]{harris2015predictive}
JR~Harris and Eric~C Grunsky.
\newblock Predictive lithological mapping of canada's north using random forest classification applied to geophysical and geochemical data.
\newblock \emph{Computers \& geosciences}, 80:\penalty0 9--25, 2015.

\bibitem[He et~al.(2016)He, Zhang, Ren, and Sun]{he2016deep}
Kaiming He, Xiangyu Zhang, Shaoqing Ren, and Jian Sun.
\newblock Deep residual learning for image recognition.
\newblock In \emph{Proceedings of the IEEE conference on computer vision and pattern recognition}, pages 770--778, 2016.

\bibitem[Kirkwood et~al.(2022)Kirkwood, Economou, Pugeault, and Odbert]{kirkwood2022bayesian}
Charlie Kirkwood, Theo Economou, Nicolas Pugeault, and Henry Odbert.
\newblock Bayesian deep learning for spatial interpolation in the presence of auxiliary information.
\newblock \emph{Mathematical Geosciences}, 54\penalty0 (3):\penalty0 507--531, 2022.

\bibitem[Maci{\k{a}}g and Harff(2020)]{macikag2020application}
{\L}ukasz Maci{\k{a}}g and Jan Harff.
\newblock Application of multivariate geostatistics for local-scale lithological mapping--case study of pelagic surface sediments from the clarion-clipperton fracture zone, north-eastern equatorial pacific (interoceanmetal claim area).
\newblock \emph{Computers \& Geosciences}, 139:\penalty0 104474, 2020.

\bibitem[{MERN, Ministère de l'Énergie et des Ressources Naturelles}(2020)]{mernchurchill2020}
{MERN, Ministère de l'Énergie et des Ressources Naturelles}.
\newblock Province de churchill, lexique stratigraphique, 2020.
\newblock URL \url{https://gq.mines.gouv.qc.ca/lexique-stratigraphique/province-de-churchill/}.

\bibitem[Murphy(2022)]{murphy2022probabilistic}
Kevin~P Murphy.
\newblock \emph{Probabilistic machine learning: an introduction}.
\newblock MIT press, 2022.

\bibitem[Oktay et~al.(2018)Oktay, Schlemper, Folgoc, Lee, Heinrich, Misawa, Mori, McDonagh, Hammerla, Kainz, et~al.]{oktay2018attention}
Ozan Oktay, Jo~Schlemper, Loic~Le Folgoc, Matthew Lee, Mattias Heinrich, Kazunari Misawa, Kensaku Mori, Steven McDonagh, Nils~Y Hammerla, Bernhard Kainz, et~al.
\newblock Attention u-net: Learning where to look for the pancreas.
\newblock \emph{arXiv preprint arXiv:1804.03999}, 2018.

\bibitem[Ronneberger et~al.(2015)Ronneberger, Fischer, and Brox]{ronneberger2015u}
Olaf Ronneberger, Philipp Fischer, and Thomas Brox.
\newblock U-net: Convolutional networks for biomedical image segmentation.
\newblock In \emph{International Conference on Medical image computing and computer-assisted intervention}, pages 234--241. Springer, 2015.

\bibitem[Shebl et~al.(2021)Shebl, Abdellatif, Hissen, Abdelaziz, and Cs{\'a}mer]{shebl2021lithological}
Ali Shebl, Mahmoud Abdellatif, Musa Hissen, Mahmoud~Ibrahim Abdelaziz, and {\'A}rp{\'a}d Cs{\'a}mer.
\newblock Lithological mapping enhancement by integrating sentinel 2 and gamma-ray data utilizing support vector machine: A case study from egypt.
\newblock \emph{International Journal of Applied Earth Observation and Geoinformation}, 105:\penalty0 102619, 2021.

\bibitem[Wang et~al.(2013)Wang, Zhang, and Li]{wang2013predictive}
Ku~Wang, Chuanrong Zhang, and Weidong Li.
\newblock Predictive mapping of soil total nitrogen at a regional scale: A comparison between geographically weighted regression and cokriging.
\newblock \emph{Applied Geography}, 42:\penalty0 73--85, 2013.

\bibitem[Wang et~al.(2021)Wang, Zuo, and Liu]{wang2021lithological}
Ziye Wang, Renguang Zuo, and Hao Liu.
\newblock Lithological mapping based on fully convolutional network and multi-source geological data.
\newblock \emph{Remote Sensing}, 13\penalty0 (23):\penalty0 4860, 2021.

\bibitem[Xia and Kulis(2017)]{xia2017w}
Xide Xia and Brian Kulis.
\newblock W-net: A deep model for fully unsupervised image segmentation.
\newblock \emph{arXiv preprint arXiv:1711.08506}, 2017.

\bibitem[Ying(2019)]{ying2019overview}
Xue Ying.
\newblock An overview of overfitting and its solutions.
\newblock In \emph{Journal of physics: Conference series}, volume 1168, page 022022. IOP Publishing, 2019.

\end{thebibliography}

\end{document}